\documentclass[conference]{IEEEtran}
\usepackage{cite}
\usepackage{amssymb,amsmath,amsthm,amsfonts}
\usepackage{algorithmic}
\usepackage[ruled,linesnumbered]{algorithm2e}
\usepackage{graphicx}
\usepackage{subcaption}
\usepackage{textcomp}
\usepackage{xcolor}
\usepackage{bm}
\usepackage{tikz}
\usepackage{url}
\usepackage{hyperref}
\newtheorem{theorem}{Theorem}
\newtheorem{definition}{Definition}
\newtheorem{example}{Example}
\newtheorem{corollary}{Corollary}

\def\BibTeX{{\rm B\kern-.05em{\sc i\kern-.025em b}\kern-.08em
    T\kern-.1667em\lower.7ex\hbox{E}\kern-.125emX}}

\newcommand{\vC}{\bm{C}}

\newcommand{\vD}{\bm{D}}

\newcommand{\vX}{\bm{X}}
\newcommand{\vS}{{\bm{S}}}

\newcommand{\czero}{\ensuremath{c_\varnothing}\xspace}

\newcommand{\singlelp}{\ensuremath{\widetilde{\mathcal{LP}}\xspace}}
\newcommand{\singledual}{\ensuremath{\widetilde{\mathcal{LD}}\xspace}}
 
\newcommand{\fulllp}{\ensuremath\mathcal{{LP}\xspace}}
\newcommand{\fulldual}{\ensuremath\mathcal{{LD}\xspace}}

\def\Bool{\ensuremath{\rm Bool}}

\SetKwFunction{InstAC}{Instrumented-AC}
\SetKwFunction{Linpassun}{LinFilter}
\SetKwFunction{Filter}{Filter}

\SetKwFunction{SimplePass}{SimplePass}
\SetKwFunction{Linpassdeux}{LinExplanation}
\SetKwFunction{Expla}{Explanation}
\SetKwFunction{Filter}{Filter}
\SetKwFunction{Tabpassdeux}{TableExplanation}
\SetKwFunction{VAClinun}{VAC-Filter}
\SetKwFunction{VAClindeux}{VAC-Tracer}
\SetKwFunction{FindLambda}{Find-Explanation$_{ia}$\&OPT}
\SetKwFunction{VAClintrois}{VAC-lin-Phase3}
\SetKwFunction{Upstruct}{Update-Counters}
\SetKwFunction{DomainC}{Bounds-Arc-Consistency}
\SetKwFunction{OPT}{Optimal-Relaxed-Solution}
\SetKwFunction{Mini}{Find-Explanation}
\SetKwFunction{extendproject}{ExtendAndProject}
\SetKwFunction{propagate}{Propagate}
\SetKwFor{Procedure}{Procedure}{}{}%
\SetKwFunction{Function}{Function}%
\SetKwFunction{Requires}{Requires}
\SetKwFunction{Proj}{Project}
\SetKwFunction{Ext}{Extend}
\SetKwFunction{Revise}{Revise}
\SetKwFunction{UProj}{UnaryProject}
\SetKwFunction{InstAC}{Instrumented-AC}
\SetKwData{Killer}{killer}
\SetKwData{True}{true}
\SetKwData{False}{false}
\SetKw{Raise}{raise}
\SetKwComment{Comment}{[}{]}
\usepackage[font={small}]{caption}
\begin{document}

\title{Virtual Arc Consistency for Linear Constraints in Cost Function Networks}

\author{\IEEEauthorblockN{Pierre Montalbano}
\IEEEauthorblockA{\textit{LIFAT, UR 6300} \\
\textit{Universit\'e de Tours, ANITI, INRAE}\\
Tours, France \\
0000-0001-8126-892X}
\and
\IEEEauthorblockN{Simon de Givry}
\IEEEauthorblockA{\textit{MIAT, UR 875} \\
\textit{Universit\'e de Toulouse, ANITI, INRAE}\\
Toulouse, France \\
0000-0002-2242-0458}
\and
\IEEEauthorblockN{George Katsirelos}
\IEEEauthorblockA{\textit{MIA Paris, AgroParisTech} \\
\textit{ANITI, INRAE}\\
Paris, France \\
0000-0002-3727-6698}
}

\maketitle

\begin{abstract}
In Constraint Programming, solving discrete minimization problems with hard and soft constraints can be done either using (i) soft global constraints, (ii) a reformulation into a linear program, or (iii) a reformulation into local cost functions. Approach (i) benefits from a vast catalog of constraints. Each soft constraint propagator communicates with other soft constraints only through the variable domains, resulting in weak lower bounds. Conversely, the approach (ii) provides a global view with strong bounds, but the size of the reformulation can be problematic. We focus on approach (iii) in which soft arc consistency (SAC) algorithms produce bounds of intermediate quality. Recently, the introduction of linear constraints as local cost functions increases their modeling expressiveness. We adapt an existing SAC algorithm to handle linear constraints. We show that our algorithm significantly improves the lower bounds compared to the original algorithm on several benchmarks, reducing solving time in some cases.\end{abstract}
%MIPLIB, CPD, PB, XCSP, and QAPLIB


\section{Introduction} 
Graphical models provide a powerful framework for modeling a variety of combinatorial problems, addressing tasks that range from satisfaction problems to probabilistic models.~\cite{cooper2020graphical}. They employ local functions defined over `small' subset of variables to represent diverse interactions among them. For example, to model the Constraint Satisfaction Problem (CSP) \cite{rossi2006handbook}, each local function is a constraint evaluating to true (satisfied) or false (falsified). Here, we focus on Cost Function Networks (CFNs), where each local function is a cost function; the task of finding an assignment that minimizes the sum of all cost functions is known as the Weighted Constraint Satisfaction Problem (WCSP). Most methods for finding optimal solutions rely on a branch-and-bound procedure, using either  
very fast but 
static and memory-intensive bounds~\cite{dechter2003mini} or memory-light ones~\cite{cooper2010soft} to compute lower bounds. Here, we focus on the latter, known as \textit{Soft Arc Consistency} (SAC) algorithms. Similar to CSP propagation, they reason about each non-unary cost function individually. Different levels of SAC exist, each offering a trade‑off between propagation strength (lower‑bound quality) and propagation time. Striking the right balance between the quality of derived lower bounds and the computational time required to construct them is crucial for achieving efficiency. \textit{Virtual Arc Consistency} (VAC)~\cite{cooper2010soft} can derive strong lower bounds but is expensive to enforce. The principle of VAC is to study a CSP, denoted $\Bool(P)$, derived from a WCSP $P$. For every cost function, only tuples and values with zero cost are allowed in $\Bool(P)$. If $\Bool(P)$ is inconsistent, then the lower bound of $P$ is non-zero. If the inconsistency of $\Bool(P)$ is detected by Generalized Arc Consistency (GAC), then VAC has been designed to extract a lower bound. %\newline

CFN also benefits from the flexibility of Constraint Programming (CP) with its ability to handle (soft)-global constraints. However, while integrating a global constraint in a CP solver only requires an algorithm to prune inconsistent values, in CFN, in addition to pruning, propagators for new constraints must also compute a lower bound. This has been done for various global constraints, including AllDifferent, clique, and linear inequality constraints \cite{allouche2016tractability,de2017clique,montalbano2022multiple}. \newline

{\bf Contributions.}
Motivated by VAC's good performance and the recent introduction of linear constraints in CFN~\cite{montalbano2022multiple}, we study how to combine these contributions. Previous approaches to handling linear constraints in CFNs tend to absorb unary costs, which subsequent propagation can no longer exploit. Enforcing VAC enables the identification of sequences of cost moves involving different propagations and facilitates communication between linear constraints.
However, enforcing VAC on a linear constraint requires keeping in $\Bool(P)$ only those values that can be part of a zero-cost tuple. For linear inequality constraints, this corresponds to solving a problem similar to the Knapsack problem and is therefore NP-hard.
We demonstrate how reduced cost filtering \cite{focacci1999cost} can be used to detect subsets of inconsistent values. This leads to VAC-lin, which enforces an incomplete GAC on $\Bool(P)$. This approach is implemented in the toulbar2 WCSP solver and tested on several benchmarks. 
\section{Background}
\subsection{Weighted Constraint Satisfaction Problem}
\begin{definition}
  A Cost Function Network (CFN) $P$ is a tuple
  $(\vX, \vD, \vC, \top)$ where $\vX = \{1, \ldots n\}$ is a set of $n$ variables,
  $\vD$ a list of finite domains $\vD_i \in \vD$ for $i \in \vX$. $\vC$ is a set of
  cost functions. 
  Each cost function $c_\vS \in \vC$ is defined over a subset of variables $\vS$ called its 
  scope $(\vS \subseteq \vX)$. $\top$
  is a maximum cost indicating a forbidden assignment.
\end{definition}
%We denote by $(i,v)$ the value $v \in \vD_i$ of variable $i \in \vX$.
%The size of the scope of a cost function is its arity.
%Unary (resp. binary) cost functions have arity 1 (resp. 2). In this paper, we assume exactly one unary cost function exists for each variable. 
%Let $\vS \subseteq \vX$ be a subset of variables, we denote by $\ell(\vS)$ the Cartesian product $\Pi_{i \in \vS} \vD_i$ of the domains of the variables in $\vS$. An assignment (or tuple) $\tau \in \ell(\vS)$ is an assignment of all the variables
%$i \in \vS$ to a value of its domain $\vD_i$. If $\vS=\vX$, then $\tau$ defines a \textit{complete assignment}; otherwise, it is a \textit{partial assignment}. We denote by $\tau[i]$ the assignment of $i \in \vS$.
%A cost function over a scope $\vS$ is denoted $c_\vS$. 
%The cost of a tuple $\tau \in \ell(\vS)$ for a cost function $c_\vS$
%is denoted $c_{\vS}(\tau)$.
%Without loss of generality, we assume all costs are positive integers, bounded
%by $\top$, a special constant signifying infeasibility. Hence if
%$c_\vS(\tau) = \top$ then the tuple $\tau$ is not feasible. In particular, setting $c_i(v)=\top$ completely forbids the assignment of $i \in \vX$ to $v \in \vD_i$. In the following, we make the unusual assumption that removed values are not excluded from the domains but forbidden by setting their unary costs to $\top$. This detail will be important in the work presented here but has no incidence on the existing SAC algorithms. 

We denote by $(i,v)$ the assignment of value $v \in \vD_i$ to variable $i \in \vX$.
The \textit{arity} of a cost function is the size of its scope.
Unary (resp. binary) cost functions have arity 1 (resp. 2). In this paper, we assume each variable has exactly one associated unary cost function. 
For $\vS \subseteq \vX$, let $\ell(\vS)=\Pi_{i \in \vS} \vD_i$ denote the Cartesian product of variable domains in $\vS$. An assignment (or tuple) $\tau \in \ell(\vS)$ is an assignment of a value in the domain $\vD_i$ to every variable
$i \in \vS$. We denote by $\tau[i]$ the value assigned to $i \in \vS$. If $\vS=\vX$, then $\tau$ defines a \textit{complete assignment}; otherwise, it is a \textit{partial assignment}. 

A cost function $c_\vS$ maps tuples $\tau \in \ell(\vS)$ to costs $c_\vS(\tau)$. The cost of a complete assignment $\tau \in \ell(\vX)$ is given by $c_P(\tau)=\sum_{c_\vS\in \vC} c_\vS(\tau[\vS])$, $\tau[\vS]$ being the projection of $\tau$ on $\vS$.
Without loss of generality, we assume all costs are positive integers, bounded
by $\top$, a special constant signifying inconsistency. If
$c_\vS(\tau) = \top$, then $\tau$ is inconsistent.
%
%In particular, setting $c_i(v)=\top$ completely forbids the assignment of $v \in \vD_i$ to $i \in \vX$. In the following, we make the unusual assumption that removed values are not excluded from the domains but forbidden by setting their unary costs to $\top$. This detail will be important in the work presented here but has no incidence on the existing SAC algorithms. 
%
A cost function $c_\vS$ is \textit{hard} if for all
$\tau \in \ell(\vS)$, $c_\vS(\tau) \in \{0,\top\}$, otherwise it is \textit{soft}. A CFN $P$
that contains only hard cost functions is a constraint network (CN).
In the following, we use the term \emph{cost function}
interchangeably with the term constraint.
The Weighted Constraint Satisfaction Problem (WCSP) 
asks, given a CFN $P$, to find a complete assignment $\tau$ minimizing
$c_P(\tau)$.  This task is NP-hard~\cite{Cooper2020}.
When the underlying CFN is a CN, the problem reduces to a CSP.
Throughout this paper, \textit{WCSP} refers to both the
optimization task and the CFN. % underlying

Each cost function is represented either in \textit{extension} or \textit{intention}. A cost function defined in extension, also known as a table constraint, explicitly lists all the tuples and their associated costs. This is feasible only for low‑arity cost functions due to memory usage that grows exponentially with arity. A cost function given in intention is defined by a function or a logical expression that specifies the relationship between the variables, for example, global constraints are typically given in intention. 

We also assume the existence of an empty scope cost function \czero representing a constant term in the objective function. Since no negative costs exist, \czero serves as a global lower bound for all assignments. It plays a key role in SAC algorithms. 
\subsection{Soft Arc Consistency}
\label{sec:SLC}
 Soft Arc Consistency (SAC) algorithms examine small subsets of cost functions sequentially. Beyond removing the locally inconsistent values, they compute a lower bound by increasing \czero. To achieve this, they rely on reparametrization: a reparameterization $P'$ of a WCSP $P$ is a WCSP with an identical structure, i.e., the set of scopes and variables is identical. The costs assigned by each individual cost function may differ, but $c_P(\tau) = c_{P'}(\tau)$ for all complete assignments $\tau$. We say that a reparametrization is better if it has a higher \czero. A reparametrization can be obtained through a set (or a sequence) of local \emph{Equivalence Preserving Transformations} (EPTs). Let
$\vS_1 \subset \vS_2$ be two scopes with corresponding cost functions
$c_{\vS_1}$ and $c_{\vS_2}$. Procedure~\ref{alg:movecost} describes
how a cost $\alpha$ moves between the corresponding cost functions.
\begin{procedure}[b]
\footnotesize
  \vspace{-0.05cm}
  \SetKwInput{KwData}{Input}
  \SetKwFunction{movecost}{MoveCost}
  \KwData{scopes $\vS_1 \subset \vS_2$}
  \KwData{$\tau_1 \in \ell(\vS_1)$}
  \KwData{cost $\alpha$ to move}
%  \BlankLine
  $c_{\vS_1}(\tau_1) \gets c_{\vS_1}(\tau_1) + \alpha$ \;
  \ForEach{$\tau_2 \in \ell(\vS_2) \mid \tau_2[{\vS_1}] = \tau_1$} {
    $c_{\vS_2}(\tau_2) \gets c_{\vS_2}(\tau_2) - \alpha$ \;
  }
  \caption{MoveCost($c_{\vS_1}, c_{\vS_2}, \tau_1, \alpha$): Move $\alpha$ units
    of cost between the tuple $\tau_1$ of scope $\vS_1$ and tuples $\tau_2$ that
    extend $\tau_1$ in scope $\vS_2$} %% (\tau_2[{\vS_1}] = \tau_1)$}
  \label{alg:movecost}
  \vspace{-0.1cm}
\end{procedure}

As a matter of terminology, when $\alpha > 0$, cost moves from the higher-arity cost function $c_{\vS_2}$ to the lower-arity $c_{\vS_1}$ and the move is called a \textit{projection}, denoted $project(c_{\vS_1},c_{\vS_2},\tau_1,\alpha)$ with $\tau_1 \in \ell(\vS_1)$.
%, equivalent to \ref{alg:movecost}$(c_{\vS_1},c_{\vS_2},\tau_1,\alpha)$.
When $\alpha < 0$, cost moves to the higher-arity cost function $c_{\vS_2}$ and the move is called an \textit{extension}, denoted $extend(c_{\vS_1},c_{\vS_2},\tau_1,-\alpha)$.%, equivalent to \ref{alg:movecost}$(c_{\vS_1},c_{\vS_2},\tau_1,\alpha)$.
When $\vS_1 = \emptyset$ and $|\vS_2| = 1$, with $\vS_2=\{i\}$, the move is called a \textit{unary projection}, denoted
$unaryProject(c_i, \alpha)$, equivalent to \ref{alg:movecost}$(\czero,c_i,\emptyset,\alpha)$. Extensions are never performed from
$\czero$, so it increases monotonically during the algorithm and along each branch of the search tree.

Identifying the cost moves that produce an optimal reparameterization—one that maximizes the lower bound—is non-trivial.
It has been shown that any
reparameterization can be derived by a set of local cost moves \cite{kolmogorov2006convergent}
and that the optimal reparameterization (with $\alpha$ rational) can be found from the optimal dual solution of a linear relaxation of the WCSP
\cite{cooper2010soft}, whose feasible region is the 
\emph{local polytope}.
%\begin{align}
%\nonumber \tag*{Local Polytope}\label{eq:localpoly} \\
%\min Obj \stackrel{\mbox{def}}{=} \czero + \sum_{\hskip -2mm i\in \vX, a\in \vD_i \hskip -2mm} c_i(a)x_{ia} + \sum_{\hskip -2mm c_\vS\in \Cplus, \tau \in\ell(\vS)\hskip -2mm}c_\vS(\tau)x_{S:\tau} \label{eq:objlocal}\\
%\textrm{s.t.} \quad \forall i \in \vX, \quad \sum_{a\in \vD_i}x_{ia} = 1\label{eq:localxi}\\
%\forall c_\vS\in \Cplus, i\in \vS, a\in \vD_i  \big(\sum_{\hskip -2mm \tau \in\ell(\vS), \tau[i]=a\hskip -2mm} x_{S:\tau}\big) - x_{ia} = 0\label{eq:localcs}
%\end{align}

However, solving this LP to optimality is often prohibitively
expensive because the worst-case time complexity of an exact LP
algorithm is $O((mr+m^2)\sqrt{m})$ \cite{Vaidya89},
where $m$ is the number of
linear constraints of the linear relaxation and $r$ their largest arity.
This poor asymptotic complexity matches empirical observation~\cite{hurley2016multi}.
Moreover, the structure of this LP does not allow more efficient algorithms, as solving LPs of this form is as hard as solving arbitrary LPs 
\cite{prusa2013universality}. Instead, research has focused on producing high-quality—though potentially suboptimal—feasible dual solutions. Various algorithms have been proposed
for this, like  Block-Coordinate Ascent (BCA) algorithms developed for image analysis \cite{kolmogorov2006convergent,Wer07,sontag08,komodakis2010mrf,sontag12,tourani2020taxonomy}
or \emph{soft arc consistencies} in constraint
programming \cite{Schiex00b,Larrosa2002,de2005existential,Zytnicki09a,cooper2010soft}.
 Notably,
the strongest algorithms from both lines of research,
such as
TRWS \cite{kolmogorov2006convergent} and VAC \cite{cooper2010soft} converge on fixpoints
with the same properties.

Here, we focus on Soft Arc Consistency (SAC)~\cite{cooper2010soft} and define several variants.
\begin{definition}
A WCSP $P$ is Node Consistent (NC)~\cite{Larrosa2002} if for every variable $i \in \vX$ there exists a value $a \in \vD_i$ such that $c_i(a)=0$ and for every value $b \in \vD_i$, $\czero + c_i(b) < \top$.
\end{definition}

In the following, we assume that the WCSP is NC before the propagator runs.

An important SAC algorithm for this paper is \textit{Virtual Arc Consistency} (VAC) \cite{cooper2010soft}. It relies on a particular CSP $\Bool(P)$ that can be derived from a WCSP instance $P$. For every cost function in $P$, except $\czero$, only the tuples and values having a zero cost are allowed in $\Bool(P)$. 
Any satisfying assignment of $\Bool(P)$ is also feasible for $P$ and, by construction, has cost $\czero$. Hence, that is an optimum assignment of $P$.
On the other hand, if $\Bool(P)$ is inconsistent, no such assignment exists, and the optimum of $P$ has a cost strictly greater than $\czero$.
It has been shown \cite{cooper2010soft} that an inconsistency certificate produced by arc consistency on $\Bool(P)$ can be used to derive a reparameterization of $P$ with increased $\czero$.
%which is not the case for stronger local consistency applied on $\Bool(P)$~\cite{Dlask23a}.

In the following, $AC(P)$ denotes the \textit{arc consistent closure} of a CSP $P$: the unique CSP obtained by removing arc‑inconsistent values from domains. An empty AC closure implies inconsistency. 

\begin{definition}[Virtual Arc Consistency~\cite{cooper2010soft}]
    A WCSP $P$ is virtual arc consistent if the (generalized) arc consistency closure of the CSP $\Bool(P)$ is non-empty.
\end{definition}

\begin{theorem}[\cite{cooper2010soft}]
\label{theorem:VAC}
Let $P$ be a WCSP such that $\czero < \top$. Then there exists a sequence of EPTs which, when applied to $P$, leads to an increase in $c_\emptyset$ if and only if the arc consistency closure of $\Bool(P)$ is empty.
\end{theorem}

The algorithm to enforce VAC is decomposed into 3 phases:

\begin{enumerate}
    \item Establish (G)AC on $\Bool(P)$. If no conflict occurred, then quit.
    \item \label{VAC-analysis}
        Given a conflict, perform \emph{conflict analysis}\footnote{This is intentionally similar
        to the term used in SAT, because it uses a post-conflict, reverse chronological order
        traversal of the operations performed during propagation.} to compute a sequence of EPTs $\bm{\sigma}$ such that applying $\bm{\sigma}$ increases $\czero$ by a cost $\lambda$.
    \item Apply $\bm{\sigma}$ to $P$ and go back to phase 1.
\end{enumerate}

To see why step \ref{VAC-analysis} is always possible, observe that arc consistency operations in $\Bool(P)$ can themselves be viewed as EPTs, where the cost moved is always $\top$. For example, pruning a value $(i,a)$ that  has lost all supports on constraint $c_{ij}$ can be viewed as extending $\top$ from each pruned value $(j,b)$ to $c_{ij}$, which
marks all supporting tuples of $(i,a)$ in $c_{ij}$ as forbidden, then projecting $\top$ from $c_{ij}$ to $(i,a)$. 
By choosing a sufficiently small  $\lambda$, we can repeat those EPTs in $P$ using $\lambda$ instead of $\top$, ensuring that no negative costs are introduced. The purpose of step \ref{VAC-analysis} is to identify
a maximal value for $\lambda$.

From the above, we see that 
as long as $\Bool(P)$ has an empty arc consistency closure, VAC will increase \czero. 

An additional heuristic variant of VAC that we consider here
is VAC$_\theta$. This uses a threshold $\theta$ when creating $\Bool_\theta(P)$ and forbids only the values/tuples with a cost greater than or equal to $\theta$. 
When $\theta = 1$, VAC$_\theta$ is equivalent to VAC.
Clearly, VAC$_\theta$ may discover only a subset of the reparameterizations that VAC can find. But the higher $\theta$ is, the higher the costs of $P$ involved in conflicts 
discovered by GAC in $\Bool_\theta(P)$. Hence, there is a chance that those lead to a higher
increase of $\czero$, although this cannot be guaranteed. On the other hand, the lower
$\theta$ is, the better the chance that $\Bool_\theta(P)$ actually has an empty AC closure.
Thus, VAC$_\theta$ is applied by starting with high values for $\theta$ to quickly
increase the lower bound and gradually decreasing it until $\theta = 1$.
In practice, a static schedule based on the original cost distribution is used~\cite{cooper2010soft}.

Cost function size strongly impacts the VAC enforcement algorithm; its time complexity is $O(ned^r)$ per iteration, where $n$ is the number of variables, $e$ the number of cost functions, $d$ the largest domain, and $r$ the largest arity. A dedicated algorithm is required to enforce a possibly weaker consistency in the presence of global constraints.
%More recently, a variable ordering heuristic exploiting information provided by VAC has been developed \cite{Trosser20a}.
%
%A stronger level than VAC can be obtained by enforcing \textit{Pair-Wise Consistency} on $\Bool(P)$ leading to \textit{Virtual Pair-Wise Consistency} \cite{montalbano2023virtual}. 
\begin{example}
    Let $P$ be a WCSP with 4 variables, each with domains $\{a,b\}$, and 5 nonzero cost functions $c_1(a)=1$, $c_4(b)=1$, $c_{12}(b,a)=1$, $c_{23}(b,a)=1$, $c_{34}(b,a)=1$, all other tuples and $\czero,c_2,c_3$ having cost 0.
    %The following tuples are associated with cost 1: $\{(1,a)\}, \{(4,b)\}, \{(1,b),(2,a)\}, \{(2,b),(3,a)\}, \{(3,b),(4,a)\}$, defining two unary (resp. three binary) cost functions $c_1$ and $c_4$ (resp. $c_{12}, c_{23}, c_{34}$) where all other tuples have cost 0.
    The AC closure of $\Bool(P)$ is empty, indeed, values $(1,a)$ and $(4,b)$ are directly removed from $\Bool(P)$ because $c_1(a)=c_4(b)>0$. Consequently, value $(2,a)$ has no support on $c_{12}$ and $(3,b)$ has no support on $c_{34}$; These values can then be removed Finally, $(2,b)$ has no support on $c_{23}$, and a domain wipe-out occurs at variable $2$. By analyzing the trace that led to this conflict, VAC deduces the following sequence of EPTs, increasing $\czero$ by one.
    
{\centering
\small	
\begin{tabular}{lll}
1) $extend(c_1,c_{12},a,1)$ & 5) $extend(c_3,c_{23},b,1)$\\
2) $extend(c_4,c_{34},b,1)$ & 6) $project(c_2,c_{23},b,1)$\\
3) $project(c_2,c_{12},a,1)$ & 7) $unaryProject(c_2,1)$\\
4) $project(c_3,c_{34},b,1)$
%% 3) $project(c_2,c_{12},a,1)$ & 7) $unaryProject(c_2,1)$ &
%% 4) $project(c_3,c_{34},b,1)$ 
\end{tabular}}

The resulting reparameterization $P'$ of $P$ is $c_{12}(a,b)=1$, $c_{23}(a,b)=1$, $c_{34}(a,b)=1$, and $\czero=1$, all other tuples and $c_1,c_2,c_3,c_4$ having cost 0.
\end{example}

\subsection{Linear Inequality Constraints}
Linear constraints are global constraints capturing a linear interaction between variables. They are expressive, compact, and used in various optimization problems, including computer science, operations research, and artificial intelligence~\cite{boros2002pseudo}. We consider a linear {\em inequality} constraint $c_\vS$ of the form: $\sum_{i \in \vS} \sum_{v \in \vD_i} w_{iv} x_{iv} \geq C$, where $c_\vS \in \vC$, $x_{iv}$ is a 0/1 variable taking value 1 when $v \in \vD_i$ is assigned to variable $i \in \vS$.
Without loss of generality, we assume the weights $w_{iv}$ and capacity $C$ are positive constants. Any linear constraint can be written in that form, using $\sum_{v \in \vD_i} x_{iv}=1$,and equality constraints can be encoded by two inequality constraints.
Initially, linear constraints are considered as hard, i.e., $\forall \tau \in \ell(\vS)$ satisfying the constraint, it holds that $c_{\vS}(\tau)=0$, otherwise $c_{\vS}(\tau)=\top$.
If EPTs involve a linear constraint, the cost of the allowed tuples can be modified, yielding  $0<c_\vS(\tau)<\top$. 

Recent work introduced a way to represent and propagate linear constraints in a WCSP solver \cite{montalbano2022multiple} through so-called \textit{delta costs}. A cost $\delta_{iv}$ is associated with each assignment $i \in \vS, v \in \vD_i$, and it captures the amount of costs moved from the unary cost function $c_i$ to a linear constraint $c_{\vS}$. A cost $\alpha > 0$ moved from $c_{\vS}$ to $c_i(v)$ decreases $\delta_{iv}$, and is denoted $project(c_i, c_{\vS}, v, \alpha) = \ref{alg:linmovecost}(c_i, c_{\vS}, v, \alpha)$. Thus, we can have negative $\delta$ costs. When $\alpha < 0$, costs move in the opposite direction and increases $\delta_{iv}$; we denote this by $extend(c_i, c_{\vS}, v, -\alpha) = \ref{alg:linmovecost}(c_i, c_{\vS}, v,\alpha)$. By $\delta_{\emptyset}$, we denote the cost moved from $c_{\vS}$ to $\czero$ ($LinProject(c_{\vS}, \alpha))$, which is necessarily positive.
Compared to \ref{alg:movecost}, \ref{alg:linmovecost} allows cost moves in constant time and space, rather than exponential in $|\vS|-1$.
\begin{procedure}[b!]
\footnotesize
  \vspace{-0.05cm}
  \SetKwInput{KwData}{Input}
  \SetKwFunction{movecost}{LinMoveCost}
  \KwData{scopes s.t. $i \in \vS$ ; $v \in \vD_i$ ; cost $\alpha$ to move}
%  \BlankLine
  $c_i(v) \gets c_i(v) + \alpha$ \;
  $\delta_{iv} \gets \delta_{iv} - \alpha$
  \caption{LinMoveCost($c_i, c_{\vS}, v, \alpha$)} %: Move $\alpha$ units of cost between $c_i(v)$ and $c_{\vS}$
  \label{alg:linmovecost}
  \vspace{-0.1cm}
\end{procedure}
\begin{procedure}[b!]
\footnotesize
  \vspace{-0.05cm}
  \SetKwInput{KwData}{Input}
  \SetKwFunction{movecost}{linProject}
  %\KwData{scopes $\vS_1 \subset \vS_2$}
  %\KwData{$\tau_1 \in \ell(\vS_1)$}
  %\KwData{cost $\alpha$ to move}
%  \BlankLine
  $\czero \gets \czero + \alpha$ \;
  $\delta_{\emptyset} \gets \delta_{\emptyset} + \alpha$
  \caption{LinProject($c_{\vS}, \alpha$)} %: Move $\alpha$ units of cost between $\czero$ and $c_{\vS}$
  \label{alg:linmovecost0}
  \vspace{-0.1cm}
\end{procedure}

%\vspace*{-0.9cm}
After any sequence of EPTs, the cost of an assignment $\tau$ is defined by:

{\small
\begin{align*}
%\label{eq:tcost}
c_{\vS}(\tau) =
    \begin{cases}
\sum_{i \in \vS}{\delta_{i\tau[i]}} - \delta_{\emptyset} & \text{if $\tau$ satisfies the constraint} \\
\top & \text{otherwise}
\end{cases}
\end{align*}
}
Initially, no cost moves have been performed, and all the $\delta$ costs are 0.  %% \newline
In \cite{montalbano2022multiple}, the authors define a SAC algorithm that moves the costs between a linear constraint $c_{\vS}$, unary cost functions $c_i, \forall i \in \vS$, and $\czero$ by solving the following linear program, denoted $\fulllp_{\vS}$, where $x_{iv}$ are relaxed to real values:

%approach the enforcement of \textit{Full} $\emptyset$\textit{-Inverse Consistency} on linear constraints.
%\begin{definition}[Full $\emptyset$-Inverse Consistency~\cite{montalbano2022multiple}]
%A WCSP is Full $\emptyset$-Inverse Consistent (F$\emptyset$IC) if for every cost function $c_\vS \in \vC$ there exists $\tau \in \ell(\vS)$ such that $c_{\vS}(\tau) + \sum_{i \in \vS} c_i(\tau[i]) = 0$.
%\end{definition}
%For a linear constraint $c_\vS$, and the associated unary cost functions,
%this can be done by solving the linear relaxation of the following $0/1$LP:
\begingroup
\small
\begin{subequations}
\begin{eqnarray}
\label{objcombined}\min z = \sum_{i \in \vS, v \in \vD_i}{(\delta_{iv}+c_i(v)) x_{iv}} - \delta_{\emptyset}
\end{eqnarray}
\begin{eqnarray}
\label{ctr}\sum_{i \in \vS, v \in \vD_i} w_{iv} x_{iv} \geq C \\
\label{AMO}\sum_{v \in \vD_i} x_{iv} = 1, & \forall i \in \vS \\
\label{bool}x_{iv}\in [0,1], & \forall i \in \vS, v \in \vD_i
\end{eqnarray}
\end{subequations}
\endgroup
This is the linear relaxation of a Multiple-Choice Knapsack Problem (MCKP), which can be solved efficiently in $O(\sum_{i \in \vS}{|\vD_i|})$~\cite{pisinger1995minimal}. 
Its dual, denoted $\fulldual_{\vS}$, is:

{\small
\begin{subequations}
\begin{eqnarray}
\label{dualobj}\max z = C y_{cc} + \sum_{i \in \vS}{y_i} - \delta_{\emptyset}\\
\label{dualctr}w_{iv} y_{cc} + y_i \leq \delta_{iv} + c_i(v), & \forall i \in \vS, v \in \vD_i\\
\label{dualpos}y_{cc} \geq 0
\end{eqnarray}
\end{subequations}
}
where $y_{cc}$ (resp. $y_i$) are real variables associated to constraints (\ref{ctr}) (resp. (\ref{AMO})) of the primal.
$\fulllp_{\vS}$ and  $\fulldual_{\vS}$ have the same optimum ({\em strong duality property}). 
From an optimal solution $z^*$, $\mathbf{x^*} = \{x^*_{iv} \mid i \in \vS, v \in \vD_i \}$ of $\fulllp_{\vS}$, we can directly deduce an optimal solution $\mathbf{y^*} = \{y^*_{cc}, y^*_i \mid i \in \vS\}$ of $\fulldual_{\vS}$. %%~\cite{montalbano2022multiple}.

%%linear program of $\fulllp_{\vS}$, it is possible to derive a set of EPTs increasing \czero by the cost of the solution~\cite{montalbano2022multiple}.

%%In the following, we will also need to find the minimal cost tuple of one linear constraint alone (without the unary cost functions). We define $\singleilp_{\vS}$ (resp $\singlelp_{\vS}$) as the $0/1$LP (resp its linear relaxation) obtained by replacing Eq.~(\ref{objcombined}) in $\fullilp_{\vS}$ by the modified objective $\min \sum_{i \in \vS, v \in \vD_i} \delta_{iv} x_{iv} - \delta_{\emptyset}$.

%%We can observe that the dual linear program of $\fulllp_{\vS}$ has the same structure as the dual of $\singlelp_{\vS}$. Any assignment (possibly inconsistent) of one dual has the same cost in the other dual. %More interestingly, 
%%Analyzing the reduced costs associated with a dual assignment $\mathbf{y}$ in problems $\fulllp_{\vS}$ and $\singlelp_{\vS}$ provides different information.

Given a dual solution $\mathbf{y^*}$, the reduced cost associated with assignment $(i,v)$ is $rc^{\mathbf{y^*}}_{\vS}(i,v) = \delta_{iv} + c_i(v) - w_{iv} y^*_{cc} - y^*_i$ and corresponds to the slack of the dual constraint $(\ref{dualctr})$. This value can be interpreted as a lower bound on the difference in the objective between any feasible solution $x$ with $x_{iv}>0$ and $\mathbf{x^*}$. We have $z-z^* \geq rc^{\mathbf{y^*}}_{\vS}(i,v)$.

In  CFNs, the reduced cost provides a lower bound on the minimal cost tuple $\tau \in \ell(\vS)$ with $\tau[i]=v$ considering only $c_\vS$ and the unary costs $c_i, \forall i \in \vS$. If $z^* > 0$ then it is possible to derive a set of EPTs, $\{extend(c_i, c_{\vS}, v, c_i(v) - rc^{\mathbf{y^*}}_{\vS}(i,v)) \mid i \in \vS, v \in \vD_i\} \bigcup \{\ref{alg:linmovecost0}(c_{\vS}, z^*)\}$, increasing \czero by $z^*$~\cite{montalbano2022multiple}. 
In the following, since we manipulate only one dual solution $\mathbf{y^*}$ at a time, we omit $\mathbf{y^*}$ and write $rc(i,v)$.

%In the context of CFN, the reduced cost computed from $\singlelp_{\vS}$ provides a lower bound on the minimal cost tuple $\tau \in \ell(\vS)$ with $\tau[i]=v$ considering only $c_\vS$, while the reduced cost from  $\fulllp_{\vS}$  accounts for both $c_\vS$ and the unary cost functions.
%These reduced costs have been exploited in  \cite{montalbano2022multiple} to enforce an incomplete version of $\fullzeroinv$ on linear constraints. However, they are not suitable for applying VAC, which primarily relies on $\Bool(P)$ where costs are restricted to $\{0,\top\}$.\\ % unary costs

\section{VAC on Linear Constraints}
A limitation of the propagation method for linear constraints introduced in \cite{montalbano2022multiple} is that constraints are propagated one-by-one and communicate only via unary cost functions. Therefore, once a linear constraint has absorbed a cost in some $\delta$, it becomes invisible to other cost functions. Moreover, the quality of the lower bound depends largely on the propagation order. Enforcing VAC$_\theta$ allows the detection of a sequence of EPTs resulting from a combination of several constraint propagations without a fixed propagation order. However, VAC$_\theta$ requires enforcing GAC on the linear constraints in $\Bool_\theta(P)$. This requires verifying for each $c_\vS \in \vC, i \in \vS, v \in \vD_i$, whether there exists a tuple $\tau \in \ell(\vS), \tau[i]=v$ such that $c_\vS(\tau) < \theta$. 
Specifically, each linear constraint $c_{\vS}$ is transformed in $\Bool_\theta(P)$ into the following hard constraint:

\begingroup
\small
\begin{align}
\label{eq:bool_tcost}
\Bool_{\theta}(c_{\vS})(\tau) =
    \begin{cases}
0 & \text{if $\tau$ satisfies the constraint $and$ } \\ & \sum_{i \in \vS}{\delta_{i\tau[i]}} - \delta_{\emptyset} < \theta \\
%\sum_{\tau[i]=v} (\delta_{iv}) - \delta_{\emptyset} & \text{if $\tau$ satisfies the constraint} \\
\top & \text{otherwise}
\end{cases}
\end{align}
\endgroup

Propagating this requires solving a Knapsack problem. % defined by $\fullilp_{\vS}$.
This problem is NP-hard, but several approaches have been studied, including dynamic programming
for enforcing GAC \cite{trick2003dynamic}, approximate filtering with a \textit{fully polynomial time approximation scheme}
\cite{sellmann2003approximated,sellmann2004practice}, and linear programming-based reduced cost filtering \cite{focacci1999cost,claus2020analysis}.

We show in the following section that we can use bounds propagation, an optimal dual solution, and reduced cost filtering \cite{focacci1999cost} to detect a subset of inconsistent tuples in $\Bool_\theta(P)$. % and to produce explanations.

\subsection{Filtering for Linear Constraints with Assignment Costs}
\label{sec:filter}

Observe that in (\ref{eq:bool_tcost}), a tuple $\tau$ is forbidden either if it violates the constraint ({\em i.e.,} the sum of the selected weights is less than the capacity), or if the delta costs are greater than or equal to the VAC threshold ($\sum_{i \in \vS}{\delta_{i\tau[i]}} - \delta_{\emptyset} \geq \theta$).
%When $\theta=\top$, only the first condition can be true.

In $\Bool_\theta(P)$, we classify the removal of a value $(i,a)$ by a hard linear constraint $\Bool_{\theta}(c_{\vS})$ as either hard or soft. A removal is \textit{hard} if no feasible tuple exists where variable $i$ takes value $a$ whatever the delta costs and $\theta$ values are, {\em i.e.},  $c_\vS(\tau)=\top, \forall \tau \in \ell(\vS) \textrm{ s.t. } \tau[i]=a$. A removal is \textit{soft} if  $\forall \tau \in \ell(\vS)  \textrm{ s.t. }  \tau[i]=a$, we have $c_\vS(\tau) \geq \theta$. Strategies for detecting and explaining hard and soft removals differ.

Hard value removals can be detected by enforcing bounds consistency on $\Bool_{\top}(c_{\vS})$, which can be performed in linear time for inequality constraints~\cite{harvey2002bounds}. 

%\textcolor{blue}{To detect a set of values that can be soft removed from  $\Bool_\theta(P)$, we solve the linear program $\fulllp_{\vS}$ for each linear constraint $c_\vS$. In $\Bool_\theta(P)$, unary cost functions are replaced by (reduced) domains. After enforcing GAC on $\Bool_{\theta}(c_i)$, we have $\Bool_{\theta}(c_i)(v) = 0, \forall v \in \vD_i$ (other values where $c_i(v) \geq \theta$ have been removed). Thus, the reduced costs obtained from solving $\fulllp_{\vS}$ and $\fulldual_{\vS}$ only depends on $\delta$ costs and allow to filter domains. Specifically, $z^* + rc(i,v)$ provides a valid lower bound on the minimal cost of any tuple $\tau \in \ell(\vS)$ with $\tau[i]=v$, accounting for $c_\vS$ and the current value removals. When $z^* + rc(i,v) \geq \theta$, $(i,v)$ can safely be removed from $\Bool_\theta(P)$. Note that this method only enables partial filtering. Achieving complete filtering would require solving a MCKP for each value, which, in practice, is too costly when filtering $\Bool_{\theta}$.} 
%

To detect a set of values that can be soft removed from $\Bool_\theta(P)$, we solve for each linear constraint $c_\vS$ a modified version of $\fulllp_{\vS}$ where $c_i(v)$ has been removed in $(\ref{objcombined})$. This modified problem is called $\singlelp_{\vS}$ and an optimal solution is denoted $\tilde{z}^*, \mathbf{\tilde{x}^*}$. In $\Bool_\theta(P)$, unary cost functions are replaced by (reduced) domains. After enforcing GAC on $\Bool_{\theta}(c_i)$, we have $\Bool_{\theta}(c_i)(v) = 0, \forall v \in \vD_i$ (other values where $c_i(v) \geq \theta$ have been removed).
The dual of $\singlelp_{\vS}$ is obtained by removing $c_i(v)$ in $(\ref{dualctr})$; we call it $\singledual_{\vS}$. The reduced costs obtained from solving $\singlelp_{\vS}$ and $\singledual_{\vS}$ allow to filter domains. Specifically, $\tilde{z}^* + rc(i,v)$ provides a valid lower bound on the minimal cost of any tuple $\tau \in \ell(\vS)$ with $\tau[i]=v$, accounting for $c_\vS$ and the current value removals. When $\tilde{z}^* + rc(i,v) \geq \theta$, $(i,v)$ can safely be removed from $\Bool_\theta(P)$.

Note that this method does not enforce complete GAC. Achieving GAC would require solving an MCKP for each value, which, in practice, is too costly when filtering $\Bool_{\theta}(P)$.

\subsection{Finding Explanations for Value Removals}
\label{sec:explain}

A further requirement of VAC is an \textit{explanation} for each value removal, primarily during phase 2. An explanation $\left\langle c_\vS,E \right\rangle$ for the removal of value $(i,a)$ is a set of values $E$ whose removal implies the removal of $(i,a)$ by arc consistency on a constraint $c_\vS$. 
An explanation is minimal if no proper subset of $E$ is an explanation.

For each hard value removal, a minimal explanation is generated using conflict explanation~\cite{hebrard2017explanation}.

For each soft removal of a value $(i,a)$ by a linear constraint $c_\vS$, we need to identify a subset of the previous value removals that are necessary to ensure that $\tilde{z}^* + rc(i,a) \geq \theta$. Let $Q$ be the list of previous value removals made by VAC; it provides a trivial non-minimal explanation.
In the second phase of VAC, we try to improve this explanation, by solving $\singlelp_{\vS}^{ia}$, a modified version of $\singlelp_{\vS}$ with additional constraints $x_{ia}=1$ and $x_{jb}=0$ for all $(j,b) \in Q$. An optimal solution is denoted $\tilde{z}^{{ia},*}, \mathbf{\tilde{x}^{{ia},*}}$. We have $\tilde{z}^{{ia},*} \geq \tilde{z}^* + rc(i,a) \geq \theta$. From this optimal primal solution, we can compute an optimal dual solution $\mathbf{\tilde{y}^{{ia},*}}$ of the dual of $\singlelp_{\vS}^{ia}$, denoted as $\singledual_{\vS}^{ia}$. Notice that constraints $x_{jb}=0$ introduce new dual variables in $\singledual_{\vS}^{ia}$ resulting in unbounded constraints ({\em i.e.,} with infinite right hand-side in Eq.$(\ref{dualctr})$). This prevents meaningful interpretation of the reduced costs associated to the values in $Q$. Instead, we choose to tighten their dual constraints by keeping $\delta_{jb}$ for the right-hand side in Eq.$(\ref{dualctr})$, as in $\singledual_{\vS}$. Thus, we compute $rc(j,b)=\delta_{jb} - w_{jb} \tilde{y}^{{ia},*}_{cc} - \tilde{y}^{{ia},*}_j$ for all $(j,b) \in Q$.
By doing so, we lose the strong duality property, but still produce valid lower bounds.

%We compute $rc(j,b)=\delta_{jb} - w_{jb} \tilde{y}^{{ia},*}_{cc} - \tilde{y}^{{ia},*}_j$ for all $(j,b) \in Q$.
%Notice that these reduced costs correspond to removed values in the primal. Thus, they should be associated to unbounded constraints in the dual ({\em i.e.,} with infinite right hand-side in Eq.$(\ref{dualctr})$), preventing meaningful interpretation. Instead, we choose to tighten their dual constraints by keeping $\delta_{iv}$ for the right-hand side in Eq.$(\ref{dualctr})$, as in $\singledual_{\vS}$.

%--------VERY  OLD VERSION------
%$z^*_{ia} \geq \theta$. While dual sensitivity analysis could theoretically identify these critical removals, the $\top$ costs assigned to removed values prevent meaningful interpretation. However, we can exploit the structural relationship between $\fulllp_{\vS}^{ia}$ and $\singlelp_{\vS}^{ia}$. 
 %Specifically, any dual assignment for $\fulllp_{\vS}^{ia}$ is also a dual assignment (possibly infeasible) for $\singlelp_{\vS}^{ia}$, and both assignments yield the same objective value. Importantly, 
 %Since $\singlelp_{\vS}^{ia}$ does not include the $\top$ unary costs, an explanation can be obtained by evaluating its associated reduced costs.
 
 %\textcolor{blue}{For a value $(j,b) \in \vR$}, the reduced cost $rc(j,b)$ represents the additional cost incurred if $x_{jb}>0$ \textcolor{blue}{ while keeping  $x_{kc}=0$, $\forall (k,c) \in \vR \setminus \{(j,b)\}$}. %Thus, $rc(j,b) + z^*_{ia}$ gives a lower bound on the cost of the minimal tuple with $x_{ia}=1$ and $x_{jb}=1$.  
Now, if $rc(j,b) \geq 0$, we can be certain that the optimal cost $\tilde{z}^{{ia},*}$ is greater than or equal to $\theta$ independently of whether $(j,b)$ is removed or not.
%$c_j(b)=0$ or $c_j(b)=\top$}. %any tuples $\tau \in \ell(\vS)$ containing both $\tau[i]=a$, $\tau[j]=b$ incurs a cost of at least $\tilde{z}^*_{ia}$. 
Consequently, $(j,b)$ cannot be part of a minimal explanation for the removal of $(i,a)$.
Otherwise, if $rc(j,b) < 0$, we include $(j,b)$ in the explanation. While this approach may result in a non-minimal explanation, finding a minimal one would require significantly more computation. Indeed, reduced costs account only the additional cost of modifying a single value and do not capture interactions when multiple values are modified together.

%\textcolor{blue}{Reduced costs provide individual lower bounds but ignore synergistic effects between variables. For example, there may exist values $(j,b),(k,c)$ such that individually $rc(j,b) + \tilde{z}^*_{ia} \geq \theta$ and  $rc(k,c) + \tilde{z}^*_{ia} \geq \theta$ but a tuple containing $x_{ia}=1$, $x_{jb}=1$, and $x_{kc}=1$ could still have a cost below $\theta$. Accounting for this is possible, but only at the cost of additional computation. Instead, we choose to compute a potentially non-minimal explanation with all values $(j,b)$ such that $rc(j,b)<0$.}

%For each removal, we only compute a straightforward explanation containing all the previously removed values, instead of a minimal subset. Obtaining a minimal explanation requires solving a new LP for each value removal, as proposed in the previous strategy. We do it only in the second phase of VAC.

%Solving $\fulllp_{\vS}$ can also help to directly detect if the linear constraint $c_\vS$ has no valid tuple in $\Bool_\theta(P)$, when $\tilde{z}^* \geq \theta$. In this case, we can also produce an explanation by analyzing the reduced costs obtained from $\singlelp_{\vS}$.

%\begin{algorithm}[H]
%\Mini{$\fulllp$}{\\
%   $\tilde{z}^* \gets$ Solve($\fulllp$) \;
%   \eIf{$\fulllp$ is infeasible}{
%   $E \gets$ Conflict Analysis \;
%   \Return{$\{\top,E\}$}
%   }{
%   Compute reduced costs from $\singlelp$ \;
%   $E \gets \{ (i,a) \mid rc(i,a) \leq 0 \}$ \;
%   \Return{$\{\tilde{z}^*,E\}$}
%   }
%}
%\caption{Function to compute absolute value}
%\end{algorithm}

\subsection{VAC-lin Subroutines}
 Here, we define VAC-lin, a local consistency obtained by enforcing in $\Bool_\theta(P)$ the filtering process of Sec.~\ref{sec:filter} on the linear constraints and GAC on the other constraints.
 Any value removed by the filtering process of Sec.~\ref{sec:filter} would have been removed by enforcing GAC on the linear constraints. Thus, the following corollary of Theorem~\ref{theorem:VAC} holds.
\begin{corollary}
\label{theorem:exist}
Let $P$ be a WCSP such that $c_\emptyset < \top$. If enforcing an incomplete GAC on the linear constraints of $\Bool_\theta(P)$, and GAC on the other constraints leads to a conflict, then there exists a sequence of EPTs which when applied to $P$ leads to an increase in $c_\emptyset$.
\end{corollary}

VAC‑lin can be enforced by integrating the techniques of Section \ref{sec:filter} into the VAC algorithm~\cite{cooper2010soft}. Algorithm \ref{alg:pass1} presents the main VAC algorithm and algorithm \ref{alg:pass2} functions specific to VAC-lin. For functions specific to table constraints, the reader can refer to \cite{cooper2010soft}. 

%More specifically, the first step requires enforcing a filtering process on $\Bool(P)$, we use domain consistency and reduced cots filtering on linear constraints and GAC on the others. This is done in function \VAClinun. If no conflict occurs then the algorithm stops, otherwise, it calls \VAClindeux which uses removal explanation to find a minimal sequence of AC operations to provoke the conflict. Then, it converts this sequence into EPTs and produces the maximum achievable increase $\lambda$ of \czero while keeping all costs non-negative. Finally, if $\lambda>0$ then the sequence of EPTs is applied to increase $\czero$. The integration of linear constraints does not change how the other constraints are propagated by VAC. Hence, we focus here on the pieces of algorithm needed to propagate linear constraints and refer the reader to \cite{cooper2010soft} to propagate the others. We also use their notation.

\subsubsection*{Filtering Phase}
The function \VAClinun in algorithm~\ref{alg:pass1} corresponds to the filtering phase, it considers $\Bool_\theta(P)$ and enforces an incomplete GAC. It ends when no more values can be removed or when a conflict appears. It uses a queue $R$ containing all constraints that require propagation. Initially, $R$ includes every constraint. Whenever a value $(i,a)$ has no support on a constraint $c_\vS$, it is removed and added to second queue $Q$. Additionally, we record the constraint responsible for this removal in a dedicated structure called $\Killer$. All constraints other than $c_\vS$ that involve the removed value $(i,a)$ must then be propagated again. Both the queue $Q$ and structure $\Killer$ are useful only for the second phase \VAClindeux.

The function $\Filter(\Bool_{\theta}(c_{\vS}))$ applies a filtering algorithm to the cost function $c_\vS$. It returns a cost and a set of variables describing either a conflict or a set of inconsistent values. Filtering of linear constraints is presented in function \Linpassun{$\Bool_{\theta}(c_{\vS})$} of Algorithm \ref{alg:pass2} and relies on the techniques described in  Section~\ref{sec:filter}.  Given a linear constraint $c_\vS$, propagation starts by enforcing bounds arc consistency and solving $\singlelp_{\vS}$. If it is conflicting (optimal cost $\geq \theta$ or inconsistent constraint), then the values involved in the conflict are identified by analysing the reduced costs, or using conflict explanation ~\cite{hebrard2017explanation}. Otherwise, if no conflict occurs, reduced costs obtained from $\singlelp_\vS$ are analyzed to remove inconsistent values. 
%To limit VAC-lin's computation time, we prioritize applying GAC to table constraints before incomplete GAC to linear constraints.
\SetAlgoNoEnd
\SetAlgoNoLine
\begin{algorithm}[hbpt]
\footnotesize
\caption{VAC general algorithm}
\label{alg:pass1}
% \tcp{ \small Propagate a linear constraint in $\Bool_\theta(P)$. 
 %Return the set of removed values or the minimum cost greater than $\theta$, and an explanation.}
%\Function{\Linpassun{$c_\vS$}}{\\

 % \If{$\tilde{z}^* \geq \theta$ \text{or $c_\vS$ is not satisfiable} \label{lineVAC1:conf1} }{
   
 %  \tcp{If $c_\vS$ is not satisfiable then we consider that $\tilde{z}^*=\top$}
 % \Return{$(\tilde{z}^*,\emptyset,E)$ \;} \label{lineVAC1:conf2}} 
 %  \label{lineVAC1:redcost} \; \tcp{Reduced cost filtering} \label{lineVAC1:redcost}
 % $E \leftarrow \{ (i,a) \mid (i,a) $ \text{has been removed in $\Bool_\theta(P)$} $\}$ \; \label{lineVAC1:trivialexp}
  %\Return{(0,Removed,E) \;}
   
%\BlankLine
\tcp{Propagate all the constraints and record the reason for each value removal. Stop when a conflict occurs or when no more values can be removed.}
\Function{\VAClinun{}}{
  \\  
   \Indp 
   $R \leftarrow \vC$ \; \label{lineVAC1:InitP}
   \While{$R\neq \varnothing$}{
    $c_\vS \leftarrow R.Pop()$ \;
    $\left\langle \tilde{z}^*, E \right\rangle \gets \Filter(\Bool_{\theta}(c_{\vS}))$ \;
    \lIf{$\tilde{z}^* \geq \theta$}{\Return{$\left\langle \tilde{z}^*,c_\vS,E \right\rangle$} }
    \ForEach{$(i,a) \in E$}{
        Delete $a$ from $\vD_i$ \;
        $\Killer(i,a) \leftarrow c_\vS$ \; \label{lineVAC1:Killer}
        $Q.Push(i,a)$ \; \label{lineVAC1:AddQ}
       \lIf{$\vD_i = \emptyset$}
        {\Return{$\left\langle \top,c_i,\{(i, b) \mid b \in \vD^{copy}_i\} \right\rangle$}} \label{lineVAC1:domainwipe}
        \lElse{$R\leftarrow R\cup \{c_{\vS'} \mid c_{\vS'}\in \vC, c_{\vS'}\neq c_\vS, i \in \vS' \}$ } \label{lineVAC1:UpdateP}
    }
   }
   \Return{$\left\langle 0,\emptyset,\emptyset \right\rangle$ \;}
   \Indm
}
\tcp{Compute $\lambda$ the maximal cost movable to \czero. 
%It relies on function \FindLambda to produce an explanation and a limiting tuple cost. Then function \Upstruct updates the structure necessary to continue the trace-back and compute a valid $\lambda$.
}

\Function{\VAClindeux{}}{ \\
 \Indp
 $Q \gets \varnothing$, $\vD^{copy} \leftarrow \vD$ \;
\label{lineVAC2:Pass1} $\left\langle \lambda,c_\vS,E \right\rangle \gets \VAClinun()$ \; 
 $\vD \leftarrow \vD^{copy}$ \;
 \label{lineVAC2:lambdalim}
 \lIf{$c_\vS \neq \emptyset$}{$k_{c_\vS} \leftarrow 1$ \label{lineVAC2:InitKctr}}
  \ForEach{$(i,a) \in E$}{$ k(i, a) \gets 1, M(i, a) \gets \True$ \;\label{lineVAC2:init}
        \lIf{$c_{i}(a) > 0$} {
          $M(i, a) \gets \False, \lambda \gets
    \min(\lambda,c_{i}(a))$ \label{lineVAC2:limitlambda}
        }
    }
 \While{$Q \neq \varnothing$}{
    $(i,a) \gets Q.Pop() $ \;
%%    $c_i(a) \gets Recover\_Unary\_Cost(i,a)$ \;
    \If{$M(i, a)$}{
      $c_\vS \gets \Killer(i,a)$ \;
        $E  \gets $\Expla{$c_\vS$ ,$(i,a)$}  \;
        \ForEach{$(j,b) \in E$}{
        $ k(j, b) \gets k(j,b)+ k(i, a)$ \; \label{lineVAC2:k}
         $ k_{c_\vS}(j, b) \leftarrow k_{c_\vS}(j, b) + k(i, a) $ \; \label{lineVAC2:kS}
          \lIf{$c_j(b) = 0$}{ $ M(j, b) \gets \True $ \label{lineVAC2:M}}
            \lElse{$\lambda \gets \min(\lambda,\frac{c_j(b)}{k(j,b)})$ }\label{lineVAC2:lambda}
	   }
      }
      }
     \Return{$\lambda$ \;} 
\Indm
}
\end{algorithm}
\subsubsection*{Tracing Phase}

Either the first phase ends with a conflict in $\Bool_\theta(P)$ or VAC-lin terminates. The purpose of \VAClindeux is to identify a minimal subset of value removals sufficient to explain this conflict. To achieve this, values involved in the conflict are marked using a Boolean function $M$. The objective is to identify a set of values/tuples with non-zero costs  that can serve as \textit{sources} to move cost to the marked values. Initially, only values returned by \VAClinun with zero unary cost are marked. Then \VAClindeux exploits the queue $Q$ and the $\Killer$ data structure to rewind the propagation history. Values are popped one by one from $Q$ and if it is marked, we trace back the cause of its deletion. \\
Identifying possible sources for a marked value $(i,a)$ can be done by computing an explanation $\left\langle c_\vS,E \right\rangle$ for its removal in $\Bool_\theta(P)$. According to Corollary \ref{theorem:exist}, moving costs from the values in $E$ to the constraint $c_\vS$ enables subsequent cost transfers from $c_\vS$ to $c_i(a)$. The constraint responsible for the removal of $(i,a)$ in  \VAClinun is stored in $\Killer(i,a)$.
Explanations are computed by function \Expla{$c_\vS,(i,a)$}. For linear constraints, this is presented in function \Linpassdeux{$c_\vS$, $(i,a)$} and relies on techniques described in \ref{sec:explain}. Specifically, the linear program $\singlelp_{\vS}^{ia}$ is solved. If it is infeasible, an explanation is generated using conflict explanation techniques~\cite{hebrard2017explanation}, otherwise, the explanation is based on reduced costs analysis. \\
Whenever values in the explanation set $E$ have zero unary cost, they are marked for further consideration. Since values in queue $Q$ are visited in reverse chronological order of their removal from $\Bool_\theta(P)$, all values in any explanation $E$ will still be present in $Q$ and will be visited later in the process. This ensures that the algorithm can systematically reconstruct a minimal explanation for the original conflict.

The maximum amount of cost $\lambda$ movable to \czero depends on the costs available at each source and the number of operations involving the marked values. Indeed, a single value can contribute to multiple removals, causing its available cost to be distributed among several cost functions. To keep track of this, we maintain three counters. The counter $k(i,a)$ records the number of request made by value $(i,a)$, while $k_{c_\vS}(i,a)$ tracks the number of request that $(i,a)$ must extend to $c_\vS$. These counters are initialized to 0 and updated each time $(i,a)$ is involved in a removal. 
Ultimately, a sequence/set of EPTs can be obtained by projecting a cost $k(i,a) \times \lambda$ from $\Killer(i,a)$ to $c_i(a)$ and extending a cost $k_{c_\vS}(i,a) \times \lambda$ from $c_i(a)$ to $c_\vS$, for all $(i,a)$ with nonzero $k$ structure. Then, a final EPT is to project a cost $\lambda$ from the conflicting constraint found by \VAClinun to $\czero$. These EPTs will be applied in VAC phase 3.
Ideally, to ensure that no negative costs are introduced, it would be useful to maintain a third counter keeping track of the number of request made by each tuple in each linear constraint. However, since the number of tuples increases exponentially with the arity of the constraint, this approach is impractical for constraints of large arity. Instead, we introduce a counter $k_{c_\vS}$ giving an upper bound on the maximal number of request made by all the tuples of $c_\vS$. This counter is only used to guarantee the correctness of $\lambda$, it is initialized to 0 and updated only when the constraint is involved in a conflict. 

The value of $\lambda$ is the largest amount of cost that can be moved while still satisfying all requests. For example, if a value $(i,a)$ has an available cost of 4  and $k(i,a)=2$, then $\lambda \leq \frac{4}{2}=2$. Initially, $\lambda$ cannot exceed the cost returned by the filtering phase; it is then as counters increase. 
\SetAlgoNoEnd
\SetAlgoNoLine 
\begin{algorithm}[hbpt]
\footnotesize
  \caption{Functions specific to VAC-lin}
  \label{alg:pass2}
%\Function{\Upstruct{$(i,a),c_\vS,KillerSet$}}{
%  \\ \ForEach{$(j,b) \in KillerSet$}{
%         $ k(j, b) \leftarrow k(j,b)+ k(i, a)$ \; \label{lineVAC2:k}
%          %$ k_{c_\vS}(j, b) \leftarrow k_{c_\vS}(j, b) + k(i, a) $ \; \label{lineVAC2:kS}
%           \lIf{$c_j(b) = 0$}{ $ M(j, b) \leftarrow \True $ \label{lineVAC2:M}}
%             \lElse{$\lambda \leftarrow \min(\lambda,\frac{c_j(b)}{k(j,b)})$}\label{lineVAC2:lambda}
% 	   }
% }
% \tcp{\small Computes an explanation for the removal of value $(i,a)$ along with an approximation of the minimal cost tuple with value $a$ assigned to variable $i$.}

% \BlankLine

\Function{\Linpassun{$\Bool_{\theta}(c_{\vS})$}}{\\
\Indp
 $Hard_{Rem} \gets$ \DomainC{$\Bool_{\top}(c_{\vS})$} \; \label{lineVAC1:BP}
 $\tilde{z}^* \leftarrow $ Solve$(\singlelp_{\vS})$ \;
 Compute reduced costs from $\singledual_{\vS}$ \;
 \If{$\singlelp_{\vS}$ is infeasible} {
    $E \leftarrow$ Conflict-explanation \; \label{lineVAC1:confexpla}
    \Return{$\left\langle \top,E \right\rangle$ \;}} \label{lineVAC1:conflict1}
 \If{$\tilde{z}^* \geq \theta$} {
    %$E \gets \{ (i,v) \mid (i,v) \in Q, i \in \vS \}$ \;
    $E \gets \{ (j,b) \in Q \mid rc(j,b) < 0 \}$ \;
\Return{$\left\langle \tilde{z}^*,E \right\rangle$ \;}} \label{lineVAC1:conflict}
 $Soft_{Rem} \leftarrow \{(i,v) \mid i \in \vS, v \in \vD_i, \tilde{z}^* + rc(i,v)  \geq \theta\}$ \; 
 \Return{$\left\langle 0, Hard_{Rem} \bigcup Soft_{Rem} \right\rangle$ \;}
\Indm}
\Function{\Linpassdeux{$c_\vS$, $(i,a)$}}{\\
\Indp
         $\tilde{z}^{{ia},*} \gets$ Solve($\singlelp_{\vS}^{ia}$) \tcc*{\small Assert: $\tilde{z}^{{ia},*} \geq \theta$}
         %with variable $i$ assigned to value $a$, and values in $Q$ removed \label{lineVAC2:findopt}\; \Comment{We have $\tilde{z}^{{ia},*} \geq \theta$}
       Compute reduced costs from $\singledual_{\vS}^{ia}$ \;
       \leIf{$\singlelp_{\vS}^{ia}$ is infeasible} { 
        $E \gets$ Conflict-explanation \; } {
        $E \gets \{ (j,b) \mid rc(j,b) < 0 \}$ } \label{lineVAC2:conflict}
      $k_{c_\vS} \leftarrow k_{c_\vS}+k(i,a)$ \;
      $\lambda \gets \min(\lambda, \frac{\tilde{z}^{{ia},*}}{k_{c_\vS}})$ \;
     \Return{$E$ \;}\label{lineVAC2:return}
    \Indm
 }

\end{algorithm}
Finally, in the third phase, all EPTs are performed according to $\Killer$, and the different $k$ structures as explained before. After this sequence of EPTs, we know a cost of $\lambda$ can be moved to $\czero$. 
Example \ref{ex:ExVAC2} illustrates one iteration of VAC-lin.

The space complexity of VAC-lin is the same as that of the original VAC~\cite{cooper2010soft}. As for time complexity, it is dominated by the filtering process. It requires solving for each constraint a relaxed knapsack problem in $O(rd)$ time~\cite{pisinger1995minimal}, where $r$ is the arity of the largest linear constraint, and $d$ is the largest domain size. The number of propagations is at most $nd$, with $n$ the number of variables. Thus, the total complexity of \VAClinun is $O(mnrd^2)$ where $m$ is the number of linear constraints. Enforcing GAC on $e'$ table constraints requires $O(e'd^{r'})$ time where $r'$ is the arity of the largest table constraints. In practice, VAC is often enforced on binary table constraints ($r'=2$), thus the complexity is often dominated by the filtering phase of the linear constraints. Therefore, GAC is enforced on table constraints before filtering linear constraints.
\begin{example}
\label{ex:ExVAC2}
    Let $P$ be a WCSP with 6 variables, with domains $\{a,b\}$, and 6 constraints $c_{12345}:7x_{1a}+7x_{2a}+3x_{3a}+3x_{4a}+3x_{5a} \geq 10$, $c_{14}: x_{1a}+x_{4b} \geq 1$, $c_{246}:x_{2b}+x_{4a}+2x_{6a} \geq 1$ and $c_{1}(a)=2$, $c_{3}(a)=2$, $c_{6}(a)=2$, all other  tuples having cost 0. Propagating the constraints as done in \cite{montalbano2022multiple} does not increase \czero. The optimal relaxed solution of this problem is $0.824$ $(\{x_{1a}=0.41176,x_{2a}=0.41176,x_{3a}=0,x_{4a}=0.41176,x_{5a}=1,x_{6a}=0\})$. We show that enforcing VAC-lin with a threshold $\theta=1$ increases \czero by 1.\newline
    In $\Bool_\theta(P)$, $(1,a)$, $(3,a)$ and $(6,a)$ are directly removed, it follows by bounds propagation on $c_{12345}$ that $(2,b)$ can be removed and we set $\Killer(2,b)=c_{12345}$. Similarly, $(4,b)$ is removed by bounds propagation on $c_{246}$ and we set $\Killer(4,b)= c_{246}$. Finally $c_{14}$ is inconsistent with explanation $\{(1,a),(4,b)\}$; thus, $\Bool_\theta(P)$ is not GAC.\newline
    We set $\lambda=\top$ and start tracing back the GAC operations. $c_{14}$ is inconsistent because $(1,a)$ and $(4,b)$ have been removed. The $k$ structures are updated: $k(1,a)=k_{c_{14}}(1,a)=k(4,b)=k_{c_{14}}(4,b)=k_{c_{14}}=1$. We immediately have $c_{1}(a)=2$. We can use this cost as a source and update $\lambda$: $\lambda = \frac{c_{1}(a)}{k(1,a)}=2$. Value $(4,b)$ verifies $c_4(b)=0$, hence, the value is marked: $M(4,b)=\True$ and must be traced. Value $(4,b)$ has been removed because it has no support on $c_{246}$, the solver computes the minimal explanation $\{(2,b),(6,a)\}$ using conflict explanation \cite{hebrard2017explanation}. The $k$ structures are updated: $k(2,b)=k_{c_{246}}(2,b)=k(6,a)=k_{c_{246}}(6,a)=k_{c_{246}}=k(4,b)=1$. We directly have $c_{6}(a)=2$, we can use this cost as a source, $\lambda$ does not need to be modified. Value $(2,b)$ verifies $c_2(b)=0$; the value is marked: $M(2,b)=\True$ and must be traced. Value $(2,b)$ has been removed because it has no support on $c_{12345}$, the solver computes the minimal explanation $\{(1,a)\}$ using conflict explanation \cite{hebrard2017explanation}.
    We update the $k$ structures: $k(1,a)=k(1,a)+k(2,b)=2$, $k_{c_{12345}}=k_{c_{12345}}(1,a)=1$. We update $\lambda$: $\lambda=\frac{c_1(a)}{k(1,a)}=1$. No more values are marked, the conflict has been explained. \newline
We deduce the following EPTs from $\Killer$, $k$ structures and $\lambda$:

{\centering
\small
\begin{tabular}{lll}
1) $extend(c_1,c_{12345},a,1)$ &  5) $project(c_4,c_{246},b,1)$ \\
2) $project(c_2,c_{12345},b,1)$ & 6) $extend(c_4,c_{14},b,1)$ \\
3) $extend(c_2,c_{246},b,1)$ &  7) $extend(c_1,c_{14},a,1)$ \\
4) $extend(c_6,c_{246},a,1)$ & 8) $\ref{alg:linmovecost0}(c_{14},1)$
%% 3) $extend(c_2,c_{246},b,1)$ & 7) $extend(c_1,c_{14},a,1)$ \\
%%4) $extend(c_6,c_{246},a,1)$ & 8) %%$project(\czero,c_{14},\emptyset,1)$ \\
\end{tabular}}

After reformulation, we have $\delta^{c_{12345}}_{1a}=\delta^{c_{246}}_{2b}=\delta^{c_{246}}_{6a}=\delta^{c_{14}}_{4b}=\delta^{c_{14}}_{1a}=\delta^{c_{14}}_{\emptyset}=1$, $\delta^{c_{12345}}_{2b}=\delta^{c_{246}}_{4b}=-1$, and $c_3(a)=2,c_6(a)=\czero=1$, all other tuples and $c_1$ having cost 0.
\end{example}

%ANOTHER POSSIBLE REFORMULATION:
%Objective: [1,7]
%Variables:
%x0 { 0 1 }/2/2 < 0 0 > s:1 Threshold: 0 (0x591851ca3c20,0) (0x591851ca2350,0)
%x1 { 0 1 }/2/2 < 0 0 > s:0 Threshold: 0 (0x591851ca4b60,0) (0x591851ca2350,1)
%x2 { 0 1 }/1/1 < 2 0 > s:1 Threshold: 0 (0x591851ca2350,2)
%x3 { 0 1 }/3/3 < 0 0 > s:0 Threshold: 0 (0x591851ca4b60,1) (0x591851ca2350,3) (0x591851ca3c20,1)
%x4 { 0 1 }/1/1 < 0 0 > s:0 Threshold: 0 (0x591851ca2350,4)
%x5 { 0 1 }/1/1 < 1 0 > s:1 Threshold: 0 (0x591851ca4b60,2)
%Constraints:
%0x591851ca2350 knapsackp(x0,x1,x2,x3,x4)  >= 10 <= 23 (ratio: 0.435) cost: 0 + 0 + (1:0:0|0:7:1,1:0:-1|0:7:0,1:0:0|0:3:0,1:0:0|0:3:0,1:0:0|0:3:0) /0.435/1,0.000,0.000,0.000,0.000,0.000 arity: 5 unassigned: 5/5/5
%0x591851ca3c20 knapsackp(x0,x3)  >= 1 <= 2 (ratio: 0.500) cost: 0 + 0 + (1:0:0|0:1:1,1:1:0|0:0:-1) /0.500/1,0.000,0.000 arity: 2 unassigned: 2/2/2
%0x591851ca4b60 knapsackp(x1,x3,x5)  >= 1 <= 3 (ratio: 0.333) cost: -1 + 0 + (1:1:1|0:0:0,1:0:0|0:1:1,1:0:0|0:1:1) /0.333/1,0.000,0.000,0.000 arity: 3 unassigned: 3/3/3

%\vspace{-0.2cm}
\section{Experimental Results}

We implemented VAC-lin in toulbar2, an open-source C++ WCSP solver.\footnote{\url{https://github.com/toulbar2/toulbar2} version 1.2.1.} The original VAC algorithm was already implemented in the solver, but only for binary cost functions in extension, with VAC maintained incrementally inside each search node~\cite{Nguyen14}). Here,
we test three variants of toulbar2. The first, a base version, applies a weaker SAC algorithm (EDAC~\cite{de2005existential} and partial F$\emptyset$IC for linear constraints~\cite{montalbano2022multiple}) at every search node of a hybrid best/depth-first branch-and-bound search method~\cite{Katsirelos15a}.
These are the default settings of toulbar2 and we denote this configuration as no-VAC. The second version, denoted VAC, is based on no-VAC and
additionally applies the existing version of VAC (which ignores linear
constraints) in preprocessing. 
The third version, denoted VAC-lin, applies during preprocessing VAC with our modifications to make it take linear constraints into account.
%Both VAC and VAC-lin are applied in preprocessing only. A weaker SAC algorithm (EDAC~\cite{de2005existential} and partial F$\emptyset$IC for linear constraints~\cite{montalbano2022multiple}) is applied at every search node of a hybrid best/depth-first branch-and-bound search method~\cite{Katsirelos15a}. We also considered solving without VAC, corresponding to default toulbar2 setting (no-VAC in the results). 
We compared these three variants of toulbar2 (no-VAC, VAC, VAC-lin) with choco, an open-source Java CP solver, and IBM cplex, a state-of-the-art integer programming solver.\footnote{\url{https://github.com/chocoteam/choco-solver} version 4.10.14 and cplex version 22.1.1.0 in single-thread mode and with non-premature stop parameters {\em EPAGAP=EPGAP=EPINT=0}.}
Choco and toulbar2 used the same {\em dom/wdeg} variable ordering heuristic~\cite{boussemart2004} with last conflict~\cite{Lecoutre09}. The value ordering heuristic is the minimum domain value for choco and EAC/VAC/VAC-lin {\em support value} for toulbar2~\cite{cooper2010soft,Trosser20a}. In VAC and VAC-lin, this corresponds to choosing first the minimum domain value in Bool(P) after doing the filtering phase. Both solvers use solution phase saving~\cite{DemirovicCP2018}.

To test our approach with a large number of linear constraints, we chose integer linear problems from the MIPLIP 2017 benchmark. We also tested the Computational Protein Design (CPD) and Quadratic Assignment Problem (QAPLIB) benchmarks, which have few additional linear constraints, large domains, and several binary cost functions in extension. We also tested a selection of the Pseudo-Boolean 2007 Evaluation benchmark (PB07). Last, to show the expressive power of CFNs with linear constraints, we experimented with the XCSP 2022/2023 benchmarks. For CPD, QAPLIB, and XCSP, we used a support encoding for cplex~\cite{hurley2016multi}. We used table constraints for choco to encode the quadratic objective of CPD and QAPLIB. 

Experiments on MIPLIB were done on a single thread of a cluster of AMD EPYC 7713 at 2.0/3.7 GHz (turbo) with 8GB and $3,600$-second CPU-time limits.
Experiments on CPD / XCSP / PB07 / QAPLIB were run on a single core of an Intel Xeon E5-2680 v3 at 2.5GHz with 64GB and $3,600$s / $2,400$ / $1,800$ / $1,200$ limits respectively.\footnote{For CPD, we add the option {\em -d:} in toulbar2 to remove its default dichotomic branching rule.}

\vspace{-0.14cm}
\subsection{MIPLIB 2017 0/1LP}

We selected 200 instances from the MIPLIB 2017 collection, containing only Boolean $0/1$ variables. Among them, 184 instances have known feasible solution.\footnote{\url{https://miplib.zib.de/tag_collection.html}}
We preprocessed them using cplex and applied our methods to the preprocessed instances.\footnote{We transformed the original real cost values in fixed-precision integer costs using 3 digits after decimal.} %% (option {\em -precision=3})
%%%{\bf WARNING! IT MAKES APPEAR SOME CONTINUOUS VARIABLES!!}
In Tab.~\ref{tab:lb_quality}, we report the average quality of lower bounds for our three variants, no-VAC, VAC, and VAC-lin, and also for the continuous linear relaxation found by cplex (column LP in Tab.~\ref{tab:lb_quality}).\footnote{The quality of an initial lower bound $l$ for a given instance with best-known solution value $b$ and trivial lower bound $t < b$ (computed as the sum of the minimum of each cost function) is defined by $(l - t)/(b-t)$. We report average quality over the number of successful instances producing a bound at the root search node for all the tested methods.} As expected, the linear relaxation provides the strongest bounds. It is also the most robust with only 5 instances where the dual simplex did not finish in 1 hour. Default toulbar2 (no-VAC) failed to produce an initial lower bound on almost 15\% of the instances, indicating that substantial engineering work remains to reach the same efficiency level as a commercial state-of-the-art LP solver. Although the original VAC algorithm is not advantageous on this benchmark due to the limited number of arity-2 linear constraints, our VAC-lin significantly improves the initial bound, going from $59\%$ to $65\%$ on average. However, it was insufficient to solve more instances for this benchmark (17 solved instances in total, whereas cplex solved 100).
%%VAC-lin solved one instance in 64.5s compared to VAC (resp. no-VAC) in 4.3s (4.7s). This instance has very large costs ({\em neos-633273}). Here, VAC-lin made 4,873 iterations in preprocessing.\footnote{This problem is already known for VAC~\cite{cooper2010soft}. Premature termination is a possible workaround.}
%%Also, we used a maximum of 3 digits of fix-point arithmetic precision to encode real cost values as integers. Reducing precision may accelerate the search.
%%Notice that toulbar2 (and choco?) are exact solvers without floating-point issues during search (internally performing integer operations only). cplex solved it in $0.17$s.
Choco performed poorly, solving 14 instances (Tab.~\ref{tab:nbsolved} and \href{https://zenodo.org/records/15691390?token=eyJhbGciOiJIUzUxMiJ9.eyJpZCI6ImQyNzk0YTUyLWRiNjUtNDdkYS1iMTM0LTFmZmYxZDk4YmU3ZSIsImRhdGEiOnt9LCJyYW5kb20iOiJmMjlhNjc4MTM5YWVhY2IyNmNhZjcxMDkzNzRmNzNiZiJ9.OnnWXxL4reecgc9FZUwczQiO4GborElk17xTfrZn_D9ny6vsR8oOPNkXDa9ueEZ_8nsLeNxjSykfxcNYYV7PTQ}{Supp. Fig.~1.a}). % \ref{fig:miplib}

\begin{table}
    \tiny
    \centering
    \begin{tabular}{|c|c|c|c|c|c|c|}
      \hline
      benchmark & total & \# av. & no-VAC & VAC & VAC-lin & LP \\ \hline
%      MIPLIB 2017 & 184   & 59.44\% (157) & 58.98\% (158) & 65.09\% (147) & \bf 84.64\% (179) \\ \hline
      MIPLIB 2017 & 184 & 147 & 59.51\% (157) & 59.23\% (158) & 65.09\% (147) & \bf 82.20\% (179) \\ \hline
%      CPD & 30 &  95.46\% (30) &  97.37\% (30) &  97.67\% ({\bf 30}) &  98.13\% (25) \\ \hline
      CPD & 30 & 25 & 95.56\% ({\bf 30}) &  97.37\% ({\bf 30}) &  97.74\% ({\bf 30}) & {\bf 98.13\%} (25) \\ \hline
      PB'2007 & 77 & 77 & 62.67\% ({\bf 77}) &  63.65\% ({\bf 77}) &  83.93\% ({\bf 77}) & \bf 86.44\% (77) \\ \hline
%      XCSP'2022 & 158 & 27.19\% (136) &  29.00\% (136) &  29.43\% ({\bf 136}) &  41.47\% (100) \\ \hline
      XCSP'2022 & 158 & 100 & 19.72\% ({\bf 136}) &  21.48\% ({\bf 136}) &  21.78\% ({\bf 136}) &  {\bf 41.47\%} (100) \\ \hline
%      XCSP'2023 & 155 & 36.97\% ({\bf 137}) &  36.62\% (137) &  38.36\% (132) &  55.90\% (93)  \\ \hline
      XCSP'2023 & 155 & 88 & 28.17\% ({\bf 137}) &  27.65\% ({\bf 137}) &  27.87\% (132) &  {\bf 53.65\%} (93) \\ \hline
      QAPLIB & 132 & 132 & 5.16\% ({\bf 132}) &  5.16\% ({\bf 132}) &  11.07\% ({\bf 132}) & \bf 12.77\% (132)\\
      \hline
    \end{tabular}
    \caption{Quality of lower bounds per benchmark averaged over the number of instances (column {\em \# av.}) where all methods produced a lower bound in space and time limits. In parentheses, number of instances where a particular method produced a lower bound.}
    \label{tab:lb_quality}
\end{table}

\begin{table}
    \tiny
    \centering
    \begin{tabular}{|c|c|c|c|c|c|c|}
      \hline
      benchmark & total & choco & cplex & toulbar2-no-VAC & toulbar2-VAC & toulbar2-VAC-lin\\ \hline
      MIPLIB 2017 & 184 & 14  &  \bf 100  &  17  &  16  &  17\\ \hline
%    1065 & 42 & 54 & 55 & 192 & 523\\
%    227 & 19 & 28 & 29 & 59 & 140\\
      CPD & 30 &  0  &  19  &  26  &  \bf 28  &  \bf 28\\ \hline
      PB'2007 & 77 & 16  & \bf 67  &  56  &  58  & \bf 67\\ \hline
      XCSP'2022 & 158 & 41  & \bf 63  &  54  &  54  &  57\\ \hline
      XCSP'2023 & 155 & 20  & \bf 39  &  17  &  17  &  22\\ \hline
      QAPLIB & 132 & 17 &  22 &  31 &  31 & \bf 32\\
      \hline
    \end{tabular}
    \caption{Number of solved instances per benchmark.}
    \label{tab:nbsolved}
\end{table}

%%%Comparison on normalized dual bound quality with and without VAC-lin on MIPLIB 2017 full benchmark (1065 instances). The value of the best-known solution normalizes each dual bound obtained at the root node. A value 1 means dual and primal (best-known solution) bounds are equal.

\vspace{-0.14cm}
\subsection{Computational Protein Design with Diversity Constraints}
As in~\cite{montalbano2022multiple}, we selected 30 CPD instances with 23-97 variables and largest domain sizes 48-194. For each instance, ten diverse solutions with a Hamming distance ten were generated using toulbar2 with a dual encoding~\cite{ruffini2021guaranted}. Next, we transformed the resulting solutions into ten linear diversity constraints and added them to our original instance.
Choco could not solve any CPD instance in 1 hour. It found solutions for half of the instances with an average distance to optimality of $0.22\%$.
VAC and VAC-lin produced almost the same results, solving 28 instances optimally. VAC-lin improved the initial lower bound found by VAC in one-third of the instances (Tab.~\ref{tab:lb_quality}). The absolute initial gap was reduced by $9.93\%$, when moving from VAC to VAC-lin. However, it did not significantly reduce the number of search nodes or solving time. 
%%, except for the number of intermediate solutions (Tab.~\ref{tab:number_sol}).
%%% ($+0.38\%$ increase) ($+1.07\%$)
A different behavior was observed between VAC and VAC-lin when applying the additional upper-bound preprocessing RASPS~\cite{Trosser20a}. VAC-rasps solved 29 and VAC-lin-rasps solved 30 instances.
Finally, no-VAC solved 26 and cplex 19 instances. %%(Tab.~\ref{tab:nbsolved}, Fig.\ref{fig:cpd}).  

\vspace{-0.14cm}
\subsection{Pseudo Boolean 2007 OPT-SMALLINT-LIN-Other}

We ran experiments on 77 instances introduced at PB 2007 Evaluation. They are unweighted Max-SAT instances with $66.3\%$ of arity-2 clauses, $25\%$ of arity-3, and the remainder from arity 4 up to $3,140$.\footnote{\url{http://www.cril.univ-artois.fr/PB07/benchs/PB07-OTHER.tar}} %10 {\em aksoy/decomp} instances also contain capacity constraints and were not solved by any solver in our experiments.} %%% note that VAC and VAC-lin will not propagate arity-3 clauses but EDAC yes.
Cplex obtained the best results, solving 67 instances within the CPU-time limit of $1,800$s; VAC‑lin also solved 67 instances but was much slower than CPLEX (\href{https://zenodo.org/records/15691390?token=eyJhbGciOiJIUzUxMiJ9.eyJpZCI6ImQyNzk0YTUyLWRiNjUtNDdkYS1iMTM0LTFmZmYxZDk4YmU3ZSIsImRhdGEiOnt9LCJyYW5kb20iOiJmMjlhNjc4MTM5YWVhY2IyNmNhZjcxMDkzNzRmNzNiZiJ9.OnnWXxL4reecgc9FZUwczQiO4GborElk17xTfrZn_D9ny6vsR8oOPNkXDa9ueEZ_8nsLeNxjSykfxcNYYV7PTQ}{Supp. Fig.~1.c}).%\footnote{{\em E.g.,} it solved instance {\em aksoy/normalized-fir08\_area\_delay} in  $1,592$ seconds. cplex solved it in 18.5 seconds. The best solver in the Max-SAT Evaluation 2023 took 34.56s to solve it (WMaxCDCL-S6-HS12).}
%%The original VAC algorithm slightly improved the baseline (no-VAC).
For this benchmark, VAC-lin clearly dominates VAC and no-VAC, which solved 58 and 56 instances respectively.\footnote{VAC and no-VAC didn't solve {\em normalized-f20c10b\_001\_area\_delay} whereas VAC-lin solved it in 25.7s and cplex in 14.5s.}
The largest instance solved by VAC-lin ({\em aksoy/normalized-fir08\_area\_delay}) has $124,856$ variables and $521,620$ clauses of maximum arity $1,023$.
%%VAC-lin has also a positive impact on the value ordering heuristic which allows to explore less intermediate solutions before reaching the optimum (Tab.~\ref{tab:number_sol}). 
Choco did not perform well, solving only 16 instances. However, it found better solutions on the unsolved {\em aksoy/decomp} instances than the other competitors.%\footnote{Average objective value of $27.6$ by choco, $29.5$ by VAC-lin, $31.2$ by no-VAC, and $31.4$ by VAC. cplex did not find a solution for {\em normalized-matrix\_5x3\_4} instance.}
%%Notice that the linear relaxation is optimal on every single linear constraint because the linear coefficients are equal to 1.

\vspace{-0.14cm}
\subsection{XCSP 2022 and 2023 MiniCOP Competition}

We restricted experiments to the mini COP category of the 2022 and 2023 XCSP competitions.\footnote{\url{https://xcsp.org/competitions}}

%%Comparing instances solved by VAC, VAC-lin, and without VAC, it turns out that one instance (BeerJugs-table-05) was solved only without VAC (finding an optimal solution of cost 37 after 53 backtracks) whereas both VAC-lin and the original VAC found the optimal solution with no backtracks but failed to prove optimality in less than 40 minutes. The initial bounds were the same for all the methods. This can be explained by the variable ordering heuristic (dom/wdeg) impacted by how solutions are found. 

Although the lower bound quality of VAC-lin is slightly better than VAC (Tab.~\ref{tab:lb_quality}), it is much higher in some particular families (XCSP22/CoinsGrid, XCSP23/Auctions) where the solving time was greatly reduced compared to VAC. Thus, VAC-lin solved slightly more instances than VAC or no-VAC (Tab.~\ref{tab:nbsolved}). It also performed better than or similar to choco depending on the benchmark.\footnote{Compared to XCSP'2023 official results, choco could not solve BeerJugs-table-07, BeerJugs-table-09, BeerJugs-table-10, Sonet-s2ring02, TravelingSalesman-015-30-00, but solved HCPizza-20-20-2-8-02 and TSPTW-n040w020-1. The different parameter settings can explain this discrepancy. We used {\em dom/wdeg} instead of {\em dom/wdeg\_cacd} and added solution phase saving.}
% (Fig.~\ref{fig:xcsp22}~and~\ref{fig:xcsp23})

The strong results obtained by cplex are not surprising. They were already observed in past MiniZinc Challenges.

\vspace{-0.14cm}
\subsection{Quadratic Assignment Problem Library}

We took 132 instances from the QAPLIB.\footnote{\url{http://coral.ise.lehigh.edu/wp-content/uploads/2014/07/qapdata.tar.gz}, excluding four instances ({\em esc128, tai150b, tai256c, tho150}) of size strictly greater than 100.}
For a problem of size $N$, we expressed the quadratic objective function as a binary Weighted CSP with $N$ variables of domain size $N$. The permutation constraint is encoded for toulbar2 and cplex as forbidden tuples for any pair of two variables (i.e., they cannot take the same value), and $N$ redundant (generalized) linear constraints of arity $N$ are added to ensure that each value is assigned to at least one variable. In Choco, it is encoded as an AllDifferent constraint.

Within the CPU-time limit of $1,200$ seconds, compared to no-VAC and to VAC, VAC-lin solved the same subset of 31 instances twice as fast (48.37s on average for VAC-lin, 80.8s for VAC and 84.5s for no-VAC) and it solved one additional instance ({\em chr25a}).\footnote{VAC-lin solved {\em chr25a} in 114sec and 142,042 search nodes. cplex didn't end in 1,200s. The best $0/1$LP approach reported in \cite{ZhangAOR2013} (R-III) solved in 274s and 5,164 nodes using a Pentium 1.7GHz and cplex 9.0.} Although it has a theoretically stronger lower bound, cplex solved only 22 instances. Choco solved 17 instances.
%% (Fig.~\ref{fig:qaplib}).

%  \begin{figure}[p]
%  	\centering
%  	\includegraphics[width=0.5\linewidth]{vac_vaclin_miplib.pdf}~\hfill~\includegraphics[width=0.5\linewidth]{lp_vaclin_miplib.pdf}
%  	\caption{Quality of lower bounds on MIPLIB 2017.}
%  	\label{fig:miplib:lbs}
% \end{figure}

%% another benefit is on value heuristics
\section{Conclusion}
Although VAC-lin improved the initial lower bound compared VAC, in many cases it was insufficient to obtain significant speedups, except on PB07, QAPLIB and on some particular categories of XCSP. Applying a stronger soft arc consistency algorithm during the search can pay off for some difficult instances \cite{montalbano2023virtual}. Testing VAC‑lin in such situations remains future work. In the future, we plan to apply the methodology developed for linear constraints to other global constraints, such as AllDifferent.

%- apply VAC-lin during search, combine with VPWC
%- VAC-lin opens the door to conflict-free learning
%- F0 / VAC-lin are soft arc consistencies that could be applied to other global constraints (alldifferent)

\section*{Acknowledgment}
Our work has benefited from the AI Interdisciplinary Institute ANITI, funded by the French %{\em Investing for the Future – } 
PIA3 program under the Grant agreement ANR-19-PI3A-0004.

\bibliographystyle{IEEEtran}
\bibliography{IEEEabrv,lipics-v2021-sample-article}

\appendix

%See the full-paper version with supplementary Fig.~1 and Tables about solution quality on \href{}{arXiv.XXX}.
%See Supplementary Fig.~1 at \href{https://zenodo.org/records/15691390?token=eyJhbGciOiJIUzUxMiJ9.eyJpZCI6ImQyNzk0YTUyLWRiNjUtNDdkYS1iMTM0LTFmZmYxZDk4YmU3ZSIsImRhdGEiOnt9LCJyYW5kb20iOiJmMjlhNjc4MTM5YWVhY2IyNmNhZjcxMDkzNzRmNzNiZiJ9.OnnWXxL4reecgc9FZUwczQiO4GborElk17xTfrZn_D9ny6vsR8oOPNkXDa9ueEZ_8nsLeNxjSykfxcNYYV7PTQ}{zenodo.org/records/15691390}.

% ~\ref{fig:cactus}
 
\begin{figure*}[htbp]
    \centering
    % First row
    \begin{subfigure}[b]{0.45\textwidth}
        \centering
        \includegraphics[width=1.2\linewidth]{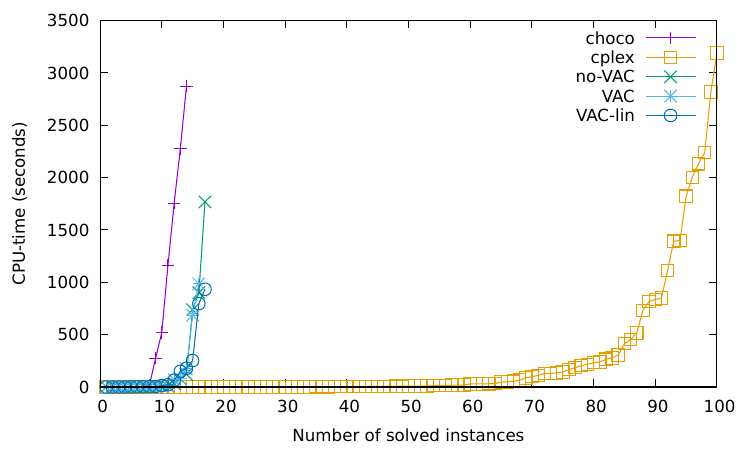}
        \caption{MIPLIB 2017}
        \label{fig:miplib}
    \end{subfigure}
    \hfill
    \begin{subfigure}[b]{0.45\textwidth}
        \centering
        \includegraphics[width=1.2\linewidth]{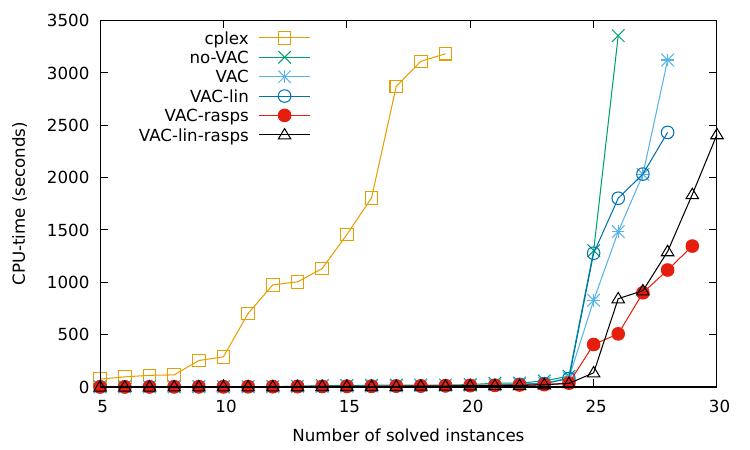}
        \caption{CPD}
        \label{fig:cpd}
    \end{subfigure}
    %\hfill    
    \vspace{-0.3\baselineskip} % Reduced vertical space
    % Second row
    \begin{subfigure}[b]{0.45\textwidth}
        \centering
        \includegraphics[width=1.2\linewidth]{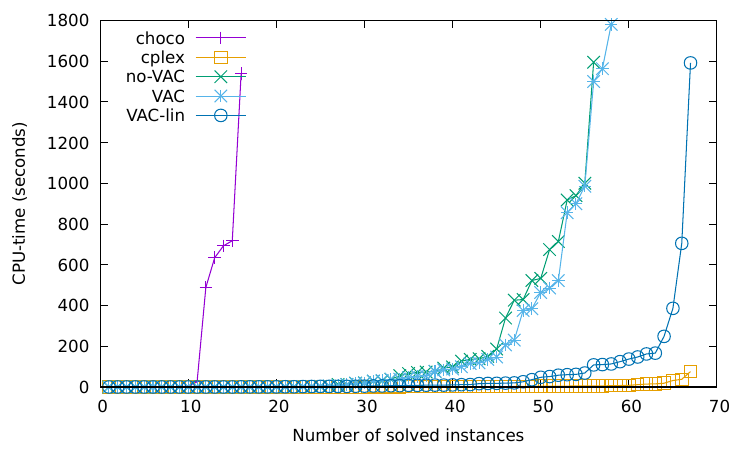}
        \caption{PB'2007}
        \label{fig:pb07}
    \end{subfigure}
    \hfill
    \begin{subfigure}[b]{0.45\textwidth}
        \centering
        \includegraphics[width=1.2\linewidth]{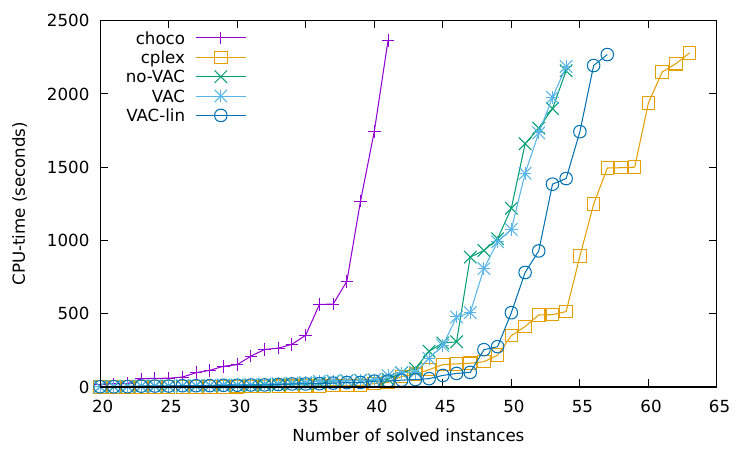}
        \caption{XCSP 2022}
        \label{fig:xcsp22}
    \end{subfigure}
    %\hfill    
    \vspace{-0.3\baselineskip} % Reduced vertical space
    % Third row (centered)
    \begin{subfigure}[b]{0.45\textwidth}
        \centering
       \includegraphics[width=1.2\linewidth]{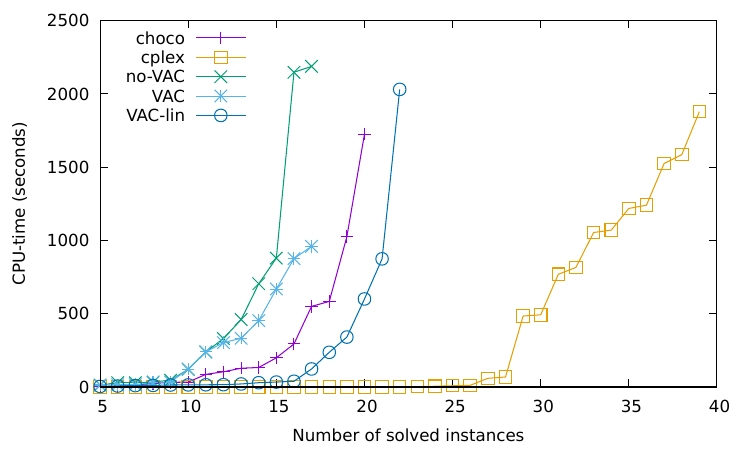}
        \caption{XCSP 2023}
        \label{fig:xcsp23}
    \end{subfigure}
    \hfill
    \begin{subfigure}[b]{0.45\textwidth}
        \centering
        \includegraphics[width=1.2\linewidth]{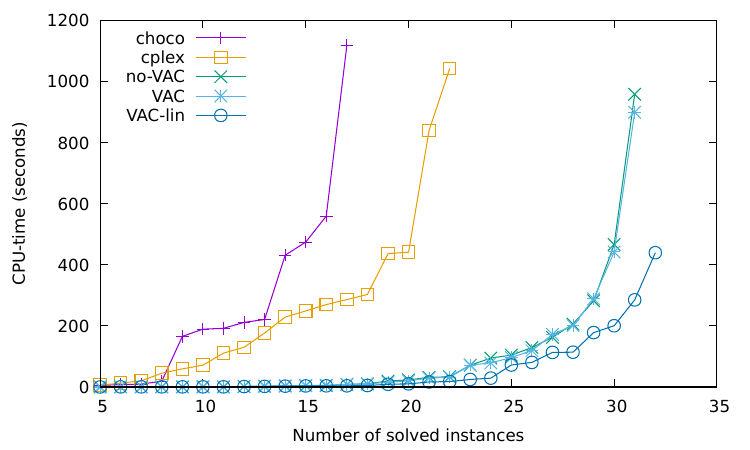}
        \caption{QAPLIB}
        \label{fig:qaplib}
    \end{subfigure}
    \caption{Supplementary figure of the paper ''Virtual Arc Consistency for Linear Constraints in Cost Function Networks'': cactus plots of CPU-time to solve various benchmarks.}
    \label{fig:cactus}
\end{figure*}

\begin{table}
    \centering
    \begin{tabular}{|c|c|c|c|c|c|c|}
      \hline
      bench & total & choco & cplex & toulbar2-no-VAC & toulbar2-VAC & toulbar2-VAC-lin \\ \hline
      MIPLIB 2017 & 184 & 244,882  ( 82 ) &  1,968 ( {\bf 164} ) &  1,720  ( 82 ) &  1,737  ( 80 ) &  1,080  ( 75 ) \\ \hline
      CPD & 30  &  788  ( 15 ) &  298 ( 25 ) &  654  ( 30 ) &  623  ( 30 ) &  \bf 577  ( 30 ) \\ \hline
      PB'2007 & 77  &  308  ( 77 ) &  315 ( 76 ) &  283  ( 77 ) &  277  ( 77 ) &  \bf 250  ( 77 ) \\ \hline
      XCSP'2022 & 158  & 10,499  ( {\bf 157} ) &  812 ( 108 ) &  2,124  ( 124 ) &  2,046  ( 124 ) &  1,829  ( 124 ) \\ \hline
      XCSP'2023 & 155  &  12,810  ( {\bf 144} ) &  491 ( 95 ) &  2,375  ( 124 ) &  2,630  ( 125 ) &  2,350  ( 112 ) \\ \hline
      QAPLIB & 132  & 5,827 (116) & 506 ( 108) & 2,214 ({\bf 132}) & 2,244 (131) & 2,200 (129)\\
      \hline
    \end{tabular}
    \caption{Total number of solutions found by each search method (in parentheses, number of instances where at least one solution has been found). E.g., on {\em XCSP22/CoinsGrid-31-14}, choco found 961 intermediate solutions before reaching the time limit, whereas no-VAC (resp. VAC) found 102 (99) intermediate solutions, VAC-lin 16 (optimality proof in 20.7s), and cplex only 1 (optimality in 0.01s).}
    \label{tab:number_sol}
\end{table}

\begin{table}
    \centering
    \begin{tabular}{|c|c|r|r|r|r|r|}
      \hline
      bench & total & choco & cplex & toulbar2-no-VAC & toulbar2-VAC & toulbar2-VAC-lin \\ \hline
      MIPLIB 2017 & 184 & 66.48\% (82) & \bf 20.61\% (164) & 40.27\% (82) & 43.00\% (80) & 39.42\% (75) \\ \hline
      CPD & 30  & 0.22\% (15) & 0.002\% (25) & 4e-6\% (30) & \bf 0.00\% (30) & \bf 0.00\% (30) \\ \hline
      PB'2007 & 77  &  41.39\% ({\bf 77}) & 0.12\% (76) & 4.65\% ({\bf 77}) & 4.65\% ({\bf 77}) & 3.09\% ({\bf 77}) \\ \hline
      XCSP'2022 & 158 & 88.49\% ({\bf 157}) & 410.76\% (108) & 13.56\% (124) & 11.89\% (124) & 11.93\% (124) \\ \hline
      XCSP'2023 & 155  & \bf 2.52\% (144) & 21.71\% (95) & 24.13\% (124) & 23.31\% (125) & 15.45\% (112) \\ \hline
      QAPLIB & 132  & 5.48\% (116) & 56.00\% (108) & 6.40\% ({\bf 132}) & 6.72\% (131) & 6.41\% (129) \\
      \hline
    \end{tabular}
    \caption{Average gap to best known solutions by each search method (in parentheses, number of instances where at least one solution has been found).}
    \label{tab:dist_to_best}
\end{table}

%QAPLIB/lipa90b instance: VAC-lin did not finished in preprocessing after 1200 seconds and 3792 iterations! When it stops, its lower bound is 1,799,624 and the optimum is 12,490,441.  In comparison, VAC took 28.7s in preprocessing, resulting in a lower bound of 1,185,639. It found a first solution of cost 15,998,465 and did not improve after that.

\end{document}